\documentclass{article}

\usepackage{microtype}
\usepackage{graphicx}
\usepackage{subcaption}
\usepackage{booktabs}
\usepackage{hyperref}
\usepackage{multirow}
\usepackage{placeins}
\usepackage{tabularx}
\usepackage{bm}
  \usepackage{enumitem}  % in preamble                                                                                                                                                       

\usepackage[preprint]{icml2026}

\usepackage{amsmath}
\usepackage{amssymb}
\usepackage{mathtools}
\usepackage{amsthm}

\usepackage[capitalize,noabbrev]{cleveref}

\theoremstyle{plain}

\theoremstyle{definition}

\theoremstyle{remark}

\icmltitlerunning{Interpreting Physics in Video World Models}

\begin{document}

% \icmlauthor[1,2,*]{Sonia Joseph}
% \icmlauthor[1]{Quentin Garrido}
% \icmlauthor[1]{Randall Balestriero}
% \icmlauthor[3]{Matthew Kowal}
% \icmlauthor[4]{Thomas Fel}
% \icmlauthor[2,\dagger]{Blake Richards}
% \icmlauthor[1,\dagger]{Mike Rabbat}

% \icmlaffiliation[1]{FAIR at Meta Superintelligence Labs}
% \icmlaffiliation[2]{McGill/Mila}
% \icmlaffiliation[3]{FAR.AI}
% \icmlaffiliation[4]{Kempner Institute, Harvard University}

% \contribution[*]{Work done at Meta}
% \contribution[\dagger]{Joint last author}
\twocolumn[
  \icmltitle{Interpreting Physics in Video World Models}

  % Symbols shown on author names
  % \icmlsetsymbol{metawork}{*}
  \icmlsetsymbol{jointlast}{\ensuremath{\dagger}}

  \begin{icmlauthorlist}

    \icmlauthor{Sonia Joseph}{meta,mila}
    \icmlauthor{Quentin Garrido}{meta}
    \icmlauthor{Randall Balestriero}{meta}
    \icmlauthor{Matthew Kowal}{farai}
    \icmlauthor{Thomas Fel}{kempner}
    \icmlauthor{Shahab Bakhtiari}{mila,montreal}

    \icmlauthor{Blake Richards}{mila,jointlast}
    \icmlauthor{Mike Rabbat}{meta,jointlast}
  \end{icmlauthorlist}

  \icmlaffiliation{meta}{FAIR, Meta Superintelligence Labs}
  \icmlaffiliation{mila}{Mila \& McGill University}
  \icmlaffiliation{farai}{FAR.AI}
  \icmlaffiliation{kempner}{Kempner Institute, Harvard University}
  \icmlaffiliation{montreal}{Université de Montréal}

  % Optional
  \icmlcorrespondingauthor{Sonia Joseph}{soniajoseph@meta.com}

  % % Notes (not footnote-style)
  % \vskip 0.08in
  % {\small
  %   % \textsuperscript{*}Work done at Meta.\quad
  %   \textsuperscript{\textdagger}Joint last authors.
  % }

  \vskip 0.3in
]

\printAffiliationsAndNotice{%
  \textsuperscript{\textdagger}Joint last authors.%
}
\begin{abstract}

A long-standing question in physical reasoning is whether video-based models need to rely on factorized representations of physical variables in order to make physically accurate predictions, or whether they can implicitly represent such variables in a task-specific, distributed manner. While modern video world models achieve strong performance on intuitive physics benchmarks, it remains unclear which of these representational regimes they implement internally. Here, we present the first interpretability study to directly examine physical representations inside large-scale video encoders. Using layerwise probing, subspace geometry, patch-level decoding, and targeted attention ablations, we characterize where physical information becomes accessible and how it is organized within encoder-based video transformers.

Across architectures, we identify a sharp intermediate-depth transition— which we call the \emph{Physics Emergence Zone}—at which physical variables become accessible. Physics-related representations peak shortly after this transition and degrade toward the output layers. Decomposing motion into explicit variables, we find that scalar quantities such as speed and acceleration are available from early layers onwards, whereas motion direction becomes accessible only at the Physics Emergence Zone. Notably, we find that direction is encoded through a high-dimensional population structure with circular geometry, requiring coordinated multi-feature intervention to control. These findings suggest that modern video models do not use factorized representations of physical variables like a classical physics engine. Instead, they use a distributed representation that is nonetheless sufficient for making physical predictions.

\end{abstract}

% ==================================================
\section{Introduction}

Despite rapid progress in video modeling, it remains unclear whether--and how--video world models represent physical information internally. Prior work has largely addressed this question indirectly, evaluating downstream performance on physics benchmarks while treating the model as a black box \citep{yi2020clevrercollisioneventsvideo,  garrido2025intuitivephysicsunderstandingemerges,motamed2025generativevideomodelsunderstand}. As a result, we lack answers to fundamental representational questions: Where in the network physical information is constructed (if at all)? How it is organized across layers and patches? What geometric form does it take?

Understanding this internal organization has implications beyond benchmark accuracy. A model that infers
stable physical structure from video could aid scientific modeling in regimes where analytic simulators are
incomplete or unavailable, including climate systems, fluid dynamics, and materials science. Models that infer physical variables in a factorized manner—i.e., with distinct representations for
quantities such as direction or momentum—could provide interpretable windows into the governing dynamics of the system being modeled
rather than black-box predictions alone. These questions closely mirror debates from cognitive science, where accounts of physics representation are
often divided between physics engine-based views, which posit compact and reusable latent state
variables \citep{battaglia2013simulation, ullman2017mind}, and heuristic-based views, which emphasize domain-specific rules or perceptual shortcuts
without an explicit physics engine \citep{siegler1976three, vasta1996water, davis2017commonsense}.

In this paper, we present one of the first interpretability analyses of physical variables in video world models. Using layerwise probing, subspace analysis, and targeted ablations, we map where physical information becomes accessible, how it is structured, and what computational substrate supports it.

\begin{table*}[h]
\centering
\caption{We consider five distinct possible representations of physics inside video world models. Our findings favor distributed, task-specific representations instead of physics-engine-style assumptions (see Section~\ref{sec:intuitive-physics-discussion} for a discussion).}
\label{tab:mechanistic_predictions}
\small
\setlength{\tabcolsep}{5pt}
\renewcommand{\arraystretch}{1.05}
\begin{tabular}{p{0.26\linewidth} p{0.34\linewidth} p{0.32\linewidth}}
\toprule
\textbf{Physics-engine assumption} &
\textbf{Interpretability prediction} &
\textbf{Our finding} \\
\midrule

Staged derivation 
& Acceleration is derived from velocity intermediates
& Acceleration is decodable at the same stage as velocity without an explicit intermediate velocity state (Sec.~\ref*{sec:cartesian-results})\\
\addlinespace[0.35em]

Cartesian representation
& Motion encoded as $(v_x, v_y)$
& Polar factorization (speed, direction) dominates (Sec.~\ref*{sec:polar-results})\\
\addlinespace[0.35em]

Shared latent physics
& Motion variables reused across physical tasks
& Direction and intuitive-physics subspaces are nearly orthogonal (Sec.~\ref*{sec:orthogonal-intphys-direction}) \\

Compact state variables
& Motion variables occupy low-dimensional subspaces
& Direction spans tens of approximately orthogonal components, steering requires dozens of dimensions (Sec.~\ref*{sec:high-feature-dimensionality})\\
\addlinespace[0.35em]

Object-centric state slots
& Motion decodable from specific spatial or temporal patches
& Direction becomes spatially redundant across patches post-Physics Emergence Zone (App.~\ref*{app:direction-patch-encoding})\\
\addlinespace[0.35em]

% Localized registers
% & Small feature sets suffice for steering
% & Individual features insufficient; population-level intervention required (Sec.~\ref*{sec:high-feature-dimensionality}) \\
\addlinespace[0.35em]

\bottomrule
\end{tabular}
\end{table*}

Physical reasoning in video world models is not formally defined, but broadly refers to the ability to infer object properties, dynamics, and interactions, such as solidity, continuity, and causality \citep{xue2023phy}. To this end, we narrow our focus to two complementary diagnostic tasks designed to examine the internal organization supporting physically coherent behavior. First, we analyze a physics task in which models distinguish possible from impossible videos \citep{riochet2021intphys, garrido2025intuitivephysicsunderstandingemerges}, capturing coarse-grained physical features including object permanence, shape constancy, and spatiotemporal continuity. Second, to enable fine-grained analysis with ground-truth physical variables, we construct a synthetic toy-ball dataset with precisely controlled velocity and acceleration. Rather than aiming for exhaustive coverage of physical reasoning, we examine in-depth how physical information is structured, transformed, and reused for these two tasks on two state-of-the-art video encoders, V-JEPA 2 (Large, Huge, and Giant) and \citep{assran2025vjepa2selfsupervisedvideo} and VideoMAE-v2 G \citep{wang2023videomaev2scalingvideo}.

Across all of the models tested, the representation of physics-related information emerges sharply at approximately one-third depth—a transition we call the \textit{Physics Emergence Zone}. The representation of physical variables then peaks in the middle layers, and degrades toward the output. Decomposing motion into finer-grained variables, we find that scalar quantities like speed and acceleration are accessible at the earliest layers, while directional information becomes accessible only at the Physics Emergence Zone. Acceleration does not require an explicit velocity intermediate and can be approximated directly by a single MLP. Direction is a reliable diagnostic of the transition, co-emerging with the ability to distinguish physically impossible videos from physically plausible ones, and with performance on temporal reasoning tasks, such as detecting shuffled videos.

We next examine the relationship between possible–impossible physics judgments and motion direction. Under a physics-engine view with compact, reusable latent states, direction would be expected to support both tasks. Instead, despite their co-emergence, direction and possible–impossible judgments occupy nearly orthogonal representational subspaces, indicating task-specific representations rather than shared latent variables. Both tasks nevertheless rely on a shared circuit-level substrate: attention heads within the Physics Emergence Zone that exhibit unusually local spatiotemporal processing. Suppressing local processing of these heads in the Physics Emergence Zone substantially degrades physics and temporal reasoning performance yet leaves static tasks such as ImageNet classification largely unaffected, showing their critical role in spatiotemporal processing. In addition, we show that motion direction is represented as a high-dimensional population code with circular geometry, reminiscent of population codes in motion neuroscience, requiring coordinated intervention across dozens of dimensions, in contrast to the low-dimensional steering in language models. Overall, our results particularly favor task-specific physical representations over compact, reusable state variables (Tab.~\ref{tab:mechanistic_predictions}).

% ==================================================
\section{Related Work}

We situate our work within prior research on video world models, physical reasoning, and interpretability of video transformers.

\subsection{Video world models}

A \emph{world model} is a learned system whose internal representations capture reusable environmental structure to support prediction, imagination, or planning \citep{ha2018worldmodels}. In visual domains, this role is increasingly fulfilled by large-scale unsupervised video models spanning video prediction, robotics, and generative modeling
\citep{ding2025understandingworldpredictingfuture,
li2025comprehensivesurveyworldmodels, kong20253d4dworldmodeling}.

While recent advances include diffusion-based generators
\citep{ho2022videodiffusionmodels, blattmann2023alignlatentshighresolutionvideo},
their computation is distributed across denoising steps, complicating representation-level analysis. We therefore focus on encoder-based video world models such as VideoMAE-v2
\citep{wang2023videomaev2scalingvideo} and V-JEPA 2 \citep{assran2025vjepa2selfsupervisedvideo},
whose persistent intermediate representations provide a natural substrate for studying internal physical structure beyond behavioral performance.

\subsection{Physical reasoning in video models and cognitive science}

Physical reasoning concerns the ability to infer object properties and interactions
such as solidity, continuity, and causality \cite{xue2023phy}. Despite strong predictive performance,
video world models exhibit systematic failures on physical reasoning benchmarks.
Models underperform on causal and counterfactual questions in CLEVRER
\citep{yi2020clevrercollisioneventsvideo}, show brittleness under violation-of-expectation
tests in IntPhys and IntPhys2
\citep{riochet2020intphysframeworkbenchmarkvisual, bordes2025intphys2benchmarkingintuitive},
and fail to generalize in interactive environments such as PHYRE
\citep{bakhtin2019phyrenewbenchmarkphysical}. Notably, perceptual realism is only
weakly correlated with physical correctness \citep{motamed2025generativevideomodelsunderstand},
raising the question of what internal representations may support physical reasoning.

In parallel, a central debate in cognitive science concerns the representational form of physical reasoning in humans: whether physical judgments rely on compact, reusable latent states—often described as an \emph{intuitive physics engine} \citep{mccloskey1983intuitive, battaglia2013simulation, ullman2017mind}—or instead arise from heuristic, domain-specific reasoning \citep{siegler1976three, vasta1996water, davis2017commonsense}. Several reviews frame this distinction as a continuum rather than a dichotomy \citep{kubricht2017intuitive, smith2023integrating}. Our work engages this debate at the representational level by analyzing how physical information is organized within video world models.

\subsection{Interpretability for video encoders}

Most previous interpretability work has focused on text and images, with comparatively limited attention to video due to its higher dimensionality and temporal complexity. Existing video interpretability studies typically rely on proxy tasks or diagnostic benchmarks to assess whether models capture dynamic information, or on visualization-based approaches such as activation maximization and clustering of intermediate representations~\cite{ghodrati2018video, hadji2018new, choi2019can, buch2022revisiting, kowal2024understanding}. These analyses have revealed strong biases toward static appearance in many video models. However, while prior work can quantify or visualize temporal concepts, it does not characterize the \textit{representational form} used to construct such concepts. In contrast, we perform targeted interpretability analyses to determine where and how physical information is organized within video encoders.

\begin{figure*}[t]
   \centering
   \includegraphics[width=0.66\textwidth]{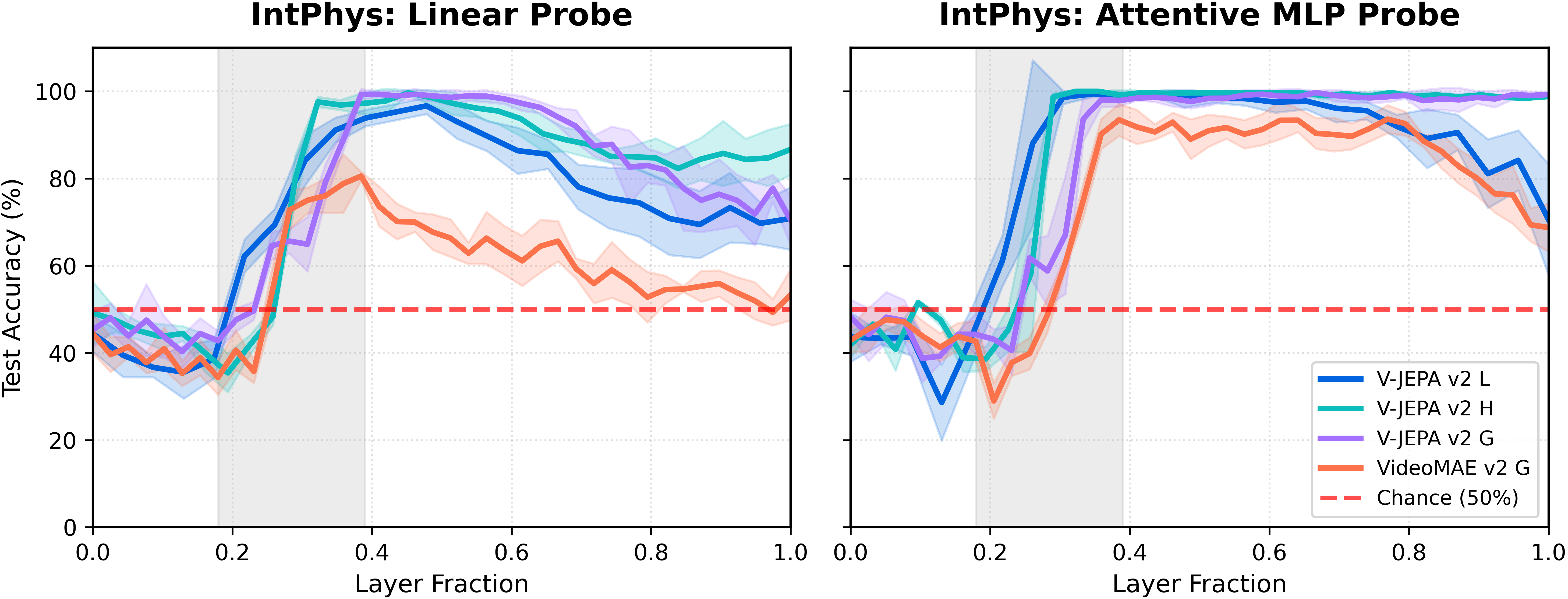}
   \caption{\textbf{The Physics Emergence Zone consistently emerges one-third in the model's layers.} We probe performance on V-JEPA 2 (Large, Huge, and Giant) and VideoMAEv2-G across all layers for the possible-vs-impossible physical reasoning task. The shaded area is the emergence zone one-third through the network, where the network starts performing well on the task for linear probes (\textit{left}) and attentive-MLP probes (\textit{right}). For full results, including for the VideoMAE-v2 family, see Appendix~\ref{app:additional_experiments:possible-vs-impossible}.}
   \label{fig:intuitive_physics_results}
\end{figure*}

\section{Models and Probing Methodology}

We study two state-of-the-art video transformer architectures and evaluate physical reasoning using a layer-wise probing methodology.

\subsection{Models}

Based on the Joint Embedding Predictive Architecture, V-JEPA 2 \citep{assran2025vjepa2selfsupervisedvideo} trains an encoder $f_{\boldsymbol{\theta}}$ to map spatiotemporal patches to latent representations, while a predictor $g_{\boldsymbol{\phi}}$ forecasts representations of masked or future patches. This predictive objective encourages temporally structured features over low-level appearance. The architecture extends a vision transformer to video using space--time patches with RoPE embeddings. We analyze the frozen pretrained encoder $f_{\boldsymbol{\theta}}$.

In contrast to V-JEPA 2's latent prediction objective, VideoMAE V2 \citep{wang2023videomaev2scalingvideo} employs masked autoencoding: a large fraction of patches is masked and an encoder-decoder reconstructs the missing pixels, with only the encoder retained after pretraining. This pixel-level reconstruction loss directly incentivizes the preservation of visual detail. We analyze the frozen pretrained encoder and contrast its internal representations with those learned by V-JEPA 2.

\subsection{Probing methodology}
% placeholder text

To localize where physical information appears in the network, we probe the residual stream at every layer. We primarily use linear probes on mean-pooled space--time patches, which provide a direct readout of what is linearly available in the representation rather than computed by the probe. Because pooling can obscure spatial or temporal structure, we complement these with patch-preserving attentive-mlp probes and interpret both jointly. This setup allows us to first identify where intuitive physics emerges, and then isolate the motion variables and mechanisms that support it. Probe training details are provided in Appendix~\ref{app:probe_training_details}.

\section{The Physics Emergence Zone}

To distinguish between physically possible and impossible sequences a model must integrate multiple perceptual cues, capturing sensitivities such as object permanence and spatiotemporal continuity \cite{baillargeon1991object, wilcox1999object, spelke1995spatiotemporal}. Although video transformers can succeed on intuitive physics benchmarks \cite{garrido2025intuitivephysicsunderstandingemerges}, behavioral performance alone does not reveal where the ability to identify impossible physical sequences emerges internally. In this section, we localize the emergence of this ability across layers and subsequently decompose it into explicit motion variables in Section~\ref{sec:motion_variables}.

\subsection{Dataset}

To test for the ability to distinguish physically possible vs. impossible sequences, we probe each layer using linear classifiers trained on the IntPhys dataset \citep{riochet2021intphys}, training the probe on a binary classification task between matched possible and impossible video pairs. Impossible variants violate core physical constraints, including object permanence (objects spontaneously appear or disappear), shape constancy (a cube transforms into a cone), or spatiotemporal continuity (trajectory reversals). Crucially, possible and impossible videos differ only at a single ``breakpoint'' frame, ensuring that successful classification requires representations that integrate high-level motion dynamics rather than visual cues like texture or color. See Appendix Fig.~\ref{app:fig:intuitive_phyiscs_dataset} for dataset examples.

\subsection{Distinction between possible and impossible physics emerges one-third through model}
\label{sec:intuitive-physics-emergence}

Across all V-JEPA 2 model scales (Large, Huge, and Giant), probe accuracies exhibit a remarkably sharp and consistent transition from near chance ($\sim$50\%) to high performance ($\sim$85--95\%) at approximately one-third of the depth through the encoder (Fig.~\ref{fig:intuitive_physics_results}). We refer to this consistent one-third depth transition as the \emph{Physics Emergence Zone}. The location of the Physics Emergence Zone is remarkably consistent across model sizes, indicating a shared computational regime rather than architecture- or scale-specific behavior. VideoMAE-v2-G also exhibits a similar depth-dependent transition; meanwhile, smaller VideoMAE-v2 variants fail to exhibit reliable emergence, potentially due to differences in model capacity, dataset size, or training objective (Appendix~\ref{app:additional_experiments:possible-vs-impossible}).

Breaking down by different types of violations of physics reveals a similar pattern across object permanence, shape constancy, and spatiotemporal continuity (Appendix Fig.~\ref{fig:intphys_data_subtask}), which indicates that the Physics Emergence Zone is not unique to a single type of violation of the laws of physics.

% \begin{figure*}[h]
%    \centering
%    \includegraphics[width=0.66\textwidth]{paper/figures/physics_variables/physics_variables.png}
%    \caption{\textbf{Direction selectively emerges at the Physics Emergence Zone}. (a) Sample frame from synthetic ball dataset, from which we measure speed and direction. (b) Results for speed, direction, and acceleration (magnitude) on V-JEPA 2-L.}
%    \label{fig:physics_variables_toy_data}
% \end{figure*} 

% \subsection{The strongest possible-vs-impossible physics representations are at the center of the network}
% \label{sec:strongest-representations-center}

In addition, perhaps counterintuitively, the best possible-vs-impossible physics representations are strongest at intermediate depth: probe accuracy peaks in the middle third of the encoder and degrades toward the output. This indicates that final-layer features do not necessarily preserve the most informative physical structure, consistent with prior findings that intermediate representations can outperform final layers for downstream perception tasks in vision encoders \cite{bolya2025perceptionencoderbestvisual}. We show that the intermediate representations of the encoder lead to better performance on a downstream intuitive task in Appendix~\ref{app:strongest-representations-downstream-task}.

% More broadly, representations learned by video world models are increasingly used as inputs to downstream systems, including physics-aware video generation pipelines \cite{yuan2025improvingphysicsvideogeneration}, motivating a closer understanding of where physically meaningful information is concentrated within the network.

\section{Velocity and Acceleration in the Physics Emergence Zone}
\label{sec:motion_variables}

\begin{figure*}[t]                                                                                                              
      \centering                                                                                                                  
      \begin{subfigure}[b]{0.32\textwidth}                                                                                        
          \centering                                                                                                              
          \includegraphics[width=\textwidth]{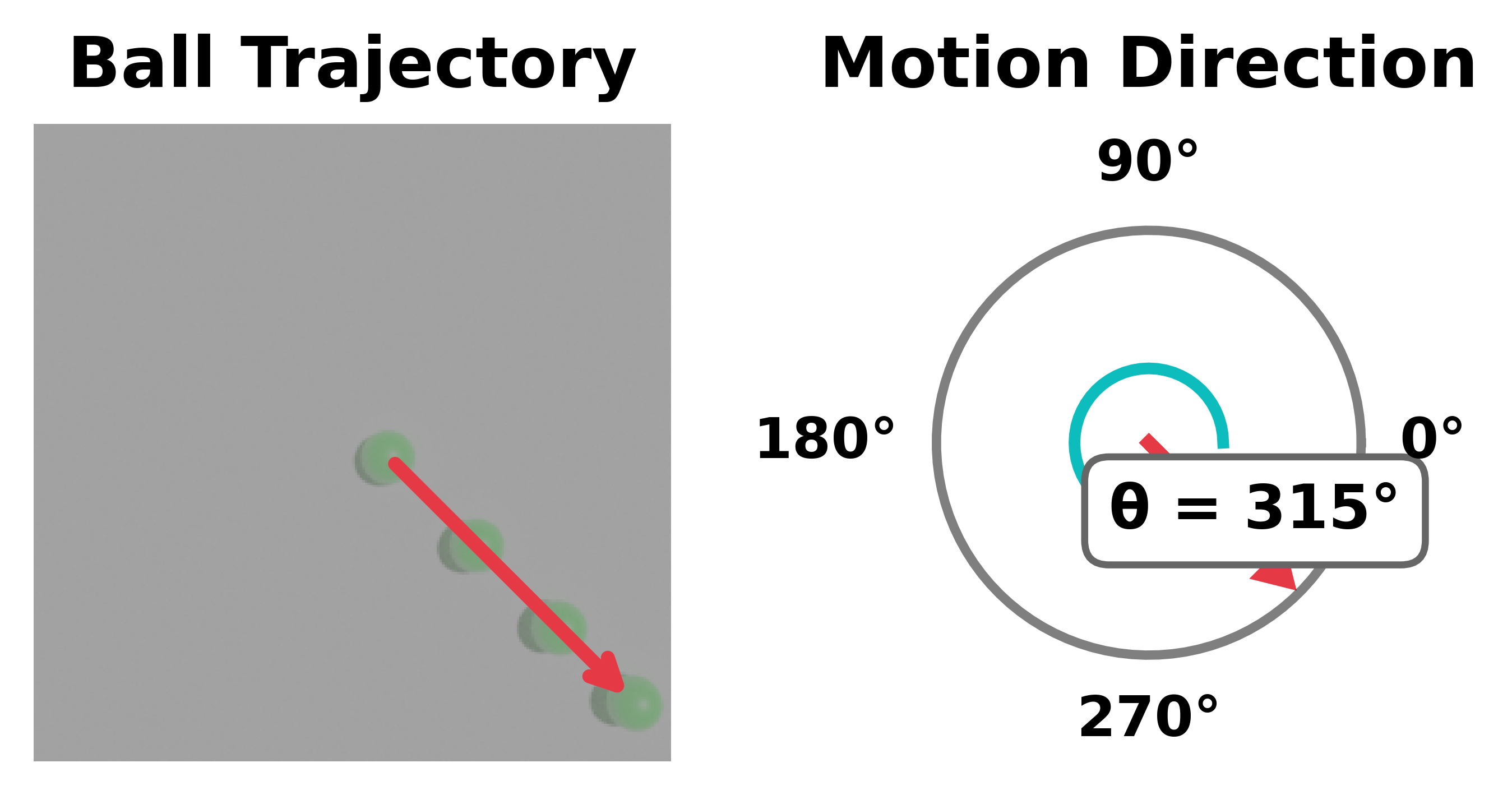}                     
          \caption{Dataset Example}                                                                                               
          \label{fig:ball_trajectory}                                                                                             
      \end{subfigure}                                                                                                             
           \hfill                                                                                                                      
      \begin{subfigure}[b]{0.32\textwidth}                                                                                        
          \centering                                                                                                              
          \includegraphics[width=\textwidth]{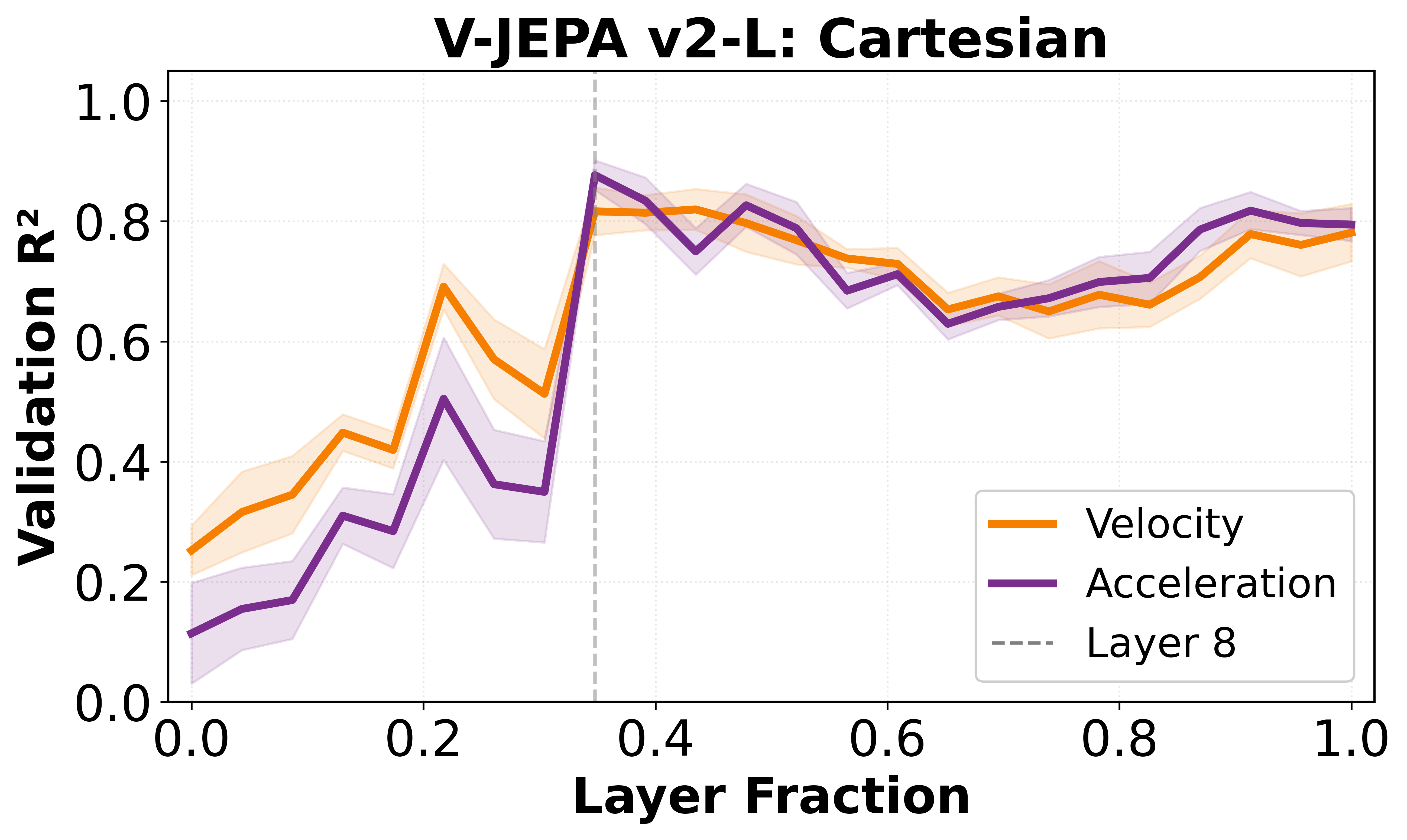}                     
          \caption{Cartesian}                                                                                                     
          \label{fig:cartesian_motion}                                                                                            
      \end{subfigure}                                                                                                             
      \hfill                                                                                                                      
      \begin{subfigure}[b]{0.32\textwidth}                                                                                        
          \centering                                                                                                              
          \includegraphics[width=\textwidth]{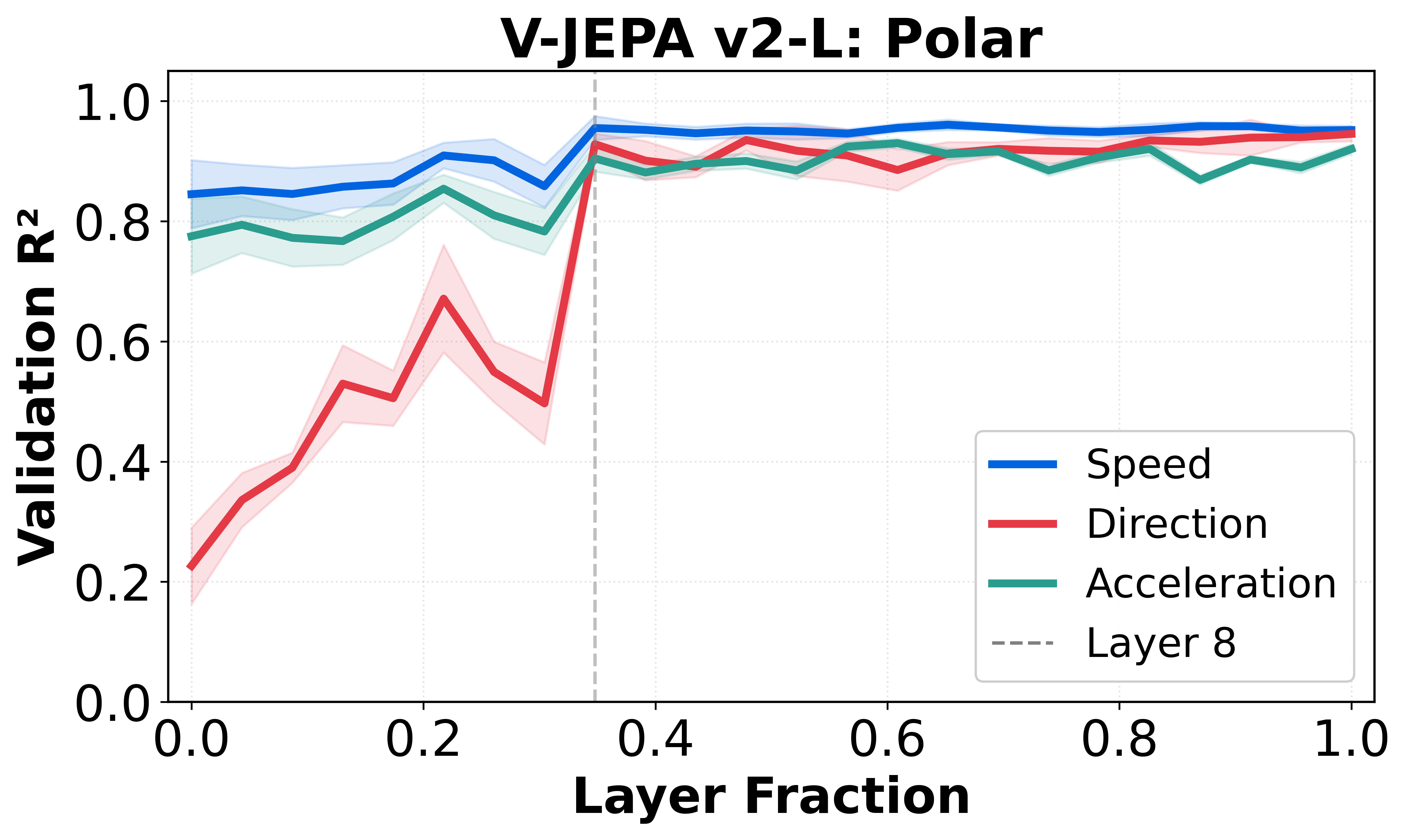}                         
          \caption{Polar}                                                                                                         
          \label{fig:polar_motion}                                                                                                
      \end{subfigure}                                                                                                             
      \caption{\textbf{Motion property encoding across layers.} (a) Example from our synthetic ball rolling dataset showing motion
   direction $\theta$. (b) Cartesian representations ($v_x$, $a_x$) show similar layer-wise emergence patterns, with acceleration available at the same time as velocity. (c) Polar         
  representations (speed, direction, acceleration magnitude) across layer fraction. Direction is only available at the Physics Emergence Zone, while magnitudes are available early.}                                                                               
      \label{fig:motion_properties}                                                                                               
  \end{figure*}     
Our results demonstrate where the models we are testing can distinguish possible from impossible physics events, but not which physical quantities contribute to that capability. To move beyond coarse behavioral signatures, we turn to explicit physical variables that admit clear ground truth. We focus on Cartesian representations for velocity $\mathbf{v}_t=(v_{x,t}, v_{y,t})$ and acceleration $\mathbf{a}_t=(a_{x,t}, a_{y,t})$, and their polar decompositions, speed $r_t=\|\mathbf{v}_t\|_2$, motion direction $\theta_t$, and acceleration magnitude $\|\mathbf{a}_t\|_2$.

\subsection{Synthetic toy ball dataset}
\label{sec:synthetic-ball-dataset}

We generate synthetic single-ball videos using the Kubric simulator \cite{greff2022kubric} with controlled motion parameters (Fig.~\ref{fig:ball_trajectory}). The ball follows straight-line trajectories under either constant velocity or externally induced acceleration, with all other factors held fixed. Ground-truth motion variables are measured in pixels per frame, enabling targeted probing of $\mathbf{v}_t$, $\mathbf{a}_t$, $r_t$, $\theta_t$, and $\|\mathbf{a}_t\|_2$. Full dataset details are provided in Appendix~\ref{app:dataset_details:toy_ball}.

\subsection{Cartesian representation: acceleration does not rely on velocity}
\label{sec:cartesian-results}

Both Cartesian velocity $(v_x, v_y)$ and acceleration $(a_x, a_y)$ exhibit a transition at the Physics Emergence Zone (Fig.~\ref{fig:cartesian_motion}). Notably, acceleration is also decodable with high $R^2$ from early layers, indicating that acceleration-like signals can be extracted directly from local features without relying on explicit intermediate velocity representations. However, Cartesian variables entangle motion magnitude and direction, making it unclear which aspects of motion are constructed at this stage.

\subsection{Polar representation: speed emerges early, while direction emerges at the Physics Emergence Zone}
\label{sec:polar-results}

To disentangle magnitude and direction, we reparameterize motion in polar coordinates. Under this decomposition, speed and acceleration magnitude are available from early layers, while directional information becomes reliably decodable only at the Physics Emergence Zone (Fig.~\ref{fig:polar_motion}). This pattern indicates that the transition is more strongly associated with the emergence of direction rather than of scalar motion quantities.  We do a deeper dive into the mechanism behind the globalization of direction in Appendix~\ref{app:direction-patch-encoding}.

The growing availability of direction in the Physics Emergence Zone persists on the multi-object dataset CLEVRER \citep{yi2020clevrercollisioneventsvideo}, in which object-level probes recover the same Physics Emergence Zone signature across object types and model scales (Appendix Fig.~\ref{fig:clevrer_r2}), indicating that direction emergence generalizes beyond single-object motion. For per-object results, see Appendix~\ref{app:additional_experiments:clevrer}.

Interestingly, the early availability of speed and the later emergence of direction mirror the motion-processing hierarchy in biological vision: speed-sensitive motion energy appears early, while higher-order pooling gives rise to position-invariant direction selectivity at later stages \cite{pasternak2020linking, born2005structure}.

% We next test whether the direction emergence persists in more complex scenes using the CLEVRER dataset \citep{yi2020clevrercollisioneventsvideo}, which contains multiple interacting objects with diverse visual attributes. We train object-level linear probes to decode motion direction at each layer. As shown in Appendix Fig.~\ref{fig:clevrer_r2}, direction decoding exhibits the same Physics Emergence Zone signature across object types and model scales. This indicates that direction emergence is not an artifact of single-object motion, but persists in multi-object environments. Per-object results are provided in Appendix~\ref{app:additional_experiments:clevrer}.

% Interestingly, the early availability of speed and the later emergence of direction are reminiscent of the motion-processing hierarchy in biological vision: simple motion energy detectors sensitive to speed appear early, while higher-order pooling gives rise to position-invariant direction selectivity at later stages. We discuss further in Section~\ref{sec:connections-to-neuroscience}

\section{What is the Relationship between Possible-vs-Impossible Physics and Direction?}

So far, we have shown that representations for possible-vs-impossible physics judgments and motion direction emerge at the same Physics Emergence Zone, but their internal relationship remains unclear. We consider three competing hypotheses about the relationship of the two representations: (i) their co-emergence reflects a generic depth-dependent effect across many tasks, including tasks not related to physical information; (ii) direction is compositionally reused to support possible-vs-impossible judgments, akin to a physics engine, or similarly, both tasks rely on the same underlying latent feature (e.g. spatiotemporal features that are more fine-grained than motion direction); or (iii) the two tasks depend on shared circuit-level computation without any representational overlap in latent space.

In this section, we evaluate each hypothesis. We find that the Physics Emergence Zone is specific to temporally structured tasks, which rules out a generic depth effect. We further show that direction and possible-vs-impossible judgments occupy distinct representational subspaces, which rules out variable reuse and shared underlying latent features. Finally, we identify a shared circuit-level substrate—local spatiotemporal processing in attention heads within the Physics Emergence Zone that supports both tasks, which provides the strongest evidence for the last hypothesis.

\subsection{The Physics Emergence Zone is specific to tasks requiring spatiotemporal processing}
\label{sec:task-specific}
We first test whether the Physics Emergence Zone reflects a generic depth-dependent pattern, or whether it is selectively associated with tasks that impose specific temporal constraints. To this end, we apply the same layerwise probing analysis to several control tasks, including CLEVRER object counting \citep{yi2020clevrercollisioneventsvideo}, ImageNet classification \citep{deng2009imagenet}, Something-Something-v2 (SSv2) video classification \citep{goyal2017somethingsomethingvideodatabase}, and shuffled versus non-shuffled video discrimination.

Although CLEVRER counting and SSv2 operate on video input, neither task necessarily requires coherent object-level motion or stable direction representations, and can in principle be supported by frame-level information or short-range temporal cues. In contrast, both IntPhys possible–impossible discrimination and shuffled-video detection impose global coherence constraints over time: IntPhys requires maintaining physically consistent object trajectories, while shuffled-video detection requires sensitivity to global temporal order.

Consistent with this distinction, neither CLEVRER counting nor SSv2 exhibits the characteristic one-third emergence signature (Appendix Fig.~\ref{fig:diverse_tasks}), whereas shuffled-video detection shows a similar emergence pattern. These results suggest that the Physics Emergence Zone is not a generic property of video processing or network depth, but is selectively associated with global spatiotemporal coherence.

\subsection{Possible-vs-impossible physics and motion direction do not overlap in latent space}
\label{sec:orthogonal-intphys-direction}

Next, we test whether the possible-vs-impossible and direction tasks share representational space: either the more general possible-vs-impossible physics task compositionally reusing direction, or both tasks relying on the same underlying feature. Both mechanisms would be similar to physics simulators, in which a general and shared set of underlying features generates a variety of physical behavior.

% We look at the geometric relationship between their decoding subspaces to find minimal overlap across model sizes: principal angles are large and cosine similarity remains low, with \cite{bjorck1973numerical}. For a full account of our method, see .  Thus, despite becoming accessible with a probe at the same depth, the two capabilities occupy nearly orthogonal representational subspaces, ruling out representational reuse and shared latent-variable explanations. 

 We look at the geometric relationship between their decoding subspaces to find minimal overlap: principal angles between motion and IntPhys subspaces average $69^\circ$–$83^\circ$, with  
  direction closer to IntPhys ($69^\circ$) than speed ($81^\circ$). Projection overlap is low—only $7$–$13\%$ of the IntPhys subspace projects onto direction, and $<3\%$ onto speed, statistically indistinguishable from random projects          
  \citep{bjorck1973numerical}. For a full account of our method and results, see Appendix~\ref{app:subspace_analysis}. Thus, despite becoming accessible with a probe at the same depth, the two capabilities occupy nearly orthogonal representational subspaces, ruling out        
  representational reuse and shared latent-variable explanations. 

% ==================================================

\subsection{Local attention heads in the Physics Emergence Zone underpin spatiotemporal processing}
\label{sec:local-head-analysis}

In the previous section, we established that the internal representations of the possible-vs-impossible physics task and direction are task-specific, without feature reuse between tasks. Still, the two tasks show a shared emergence pattern in the Physics Emergence Zone, which suggests that they may share an underlying computational process-- for example, the same attention heads or MLP mechanisms may be responsible for both tasks. Our results from Section \ref{sec:task-specific} suggest that the Physics Emergence Zone contains a mechanism that is unique to spatiotemporal processing.

Given that attention heads mediate spatiotemporal processing in transformers across patches, we analyze attention head distance across layers. We find that attention heads outside of the Physics Emergence Zone show relatively homogenous attention profiles. However, uniquely at the Physics Emergence Zone, unusually spatiotemporally local attention heads emerge alongside longer-range heads, resulting in a sharp increase in attention head diversity as measured by distance (Fig.~\ref{fig:attention_analysis}). Our method of measuring attention head distance is in Appendix~\ref{app:appendix_attention_distance}.

  \begin{figure}[h]                                                                                                               
      \centering                                                                                                                  
      \includegraphics[width=.88\columnwidth]{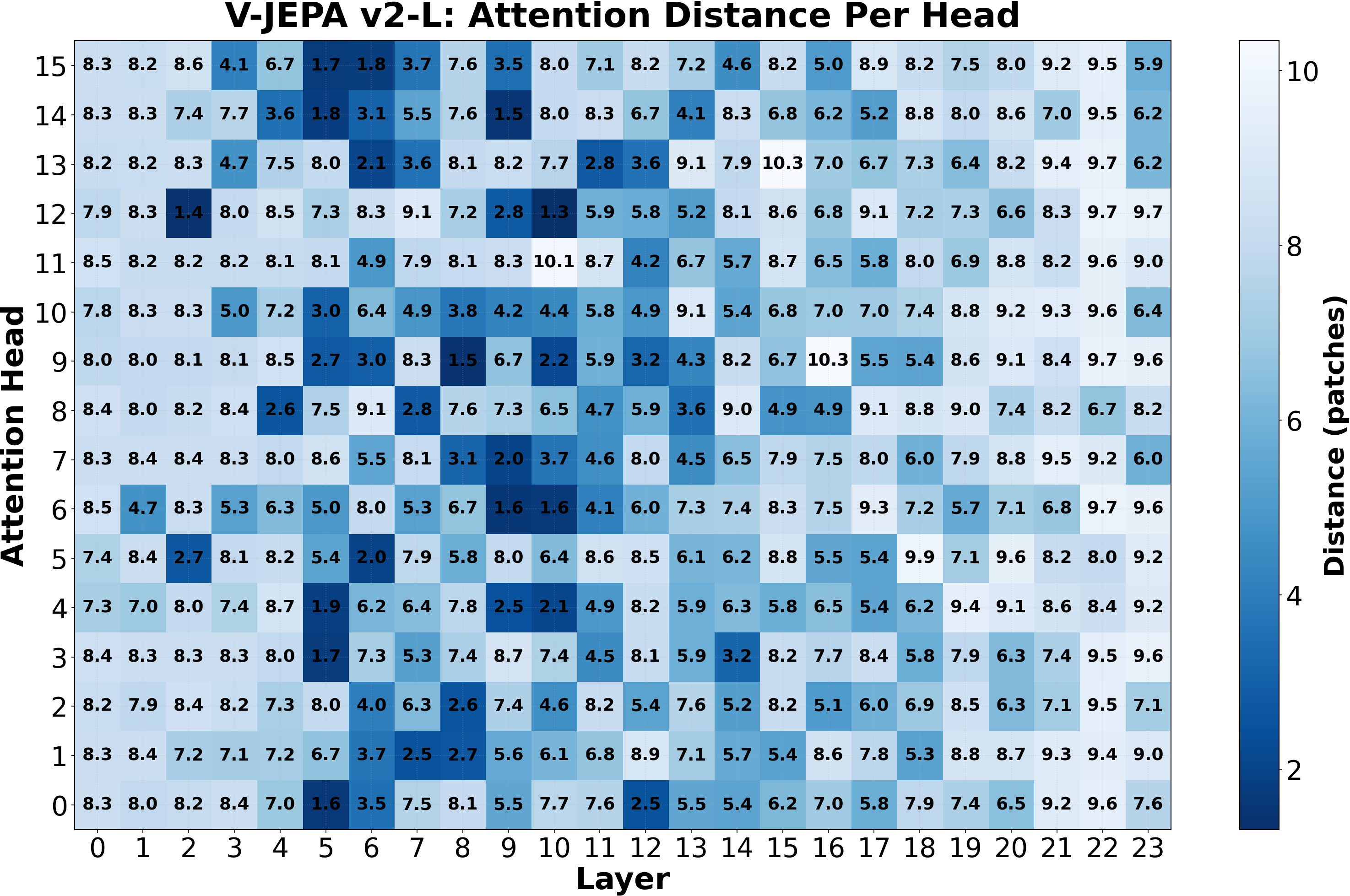}                          
      \caption{\textbf{Spatiotemporally local attention heads crop up uniquely at the Physics Emergence Zone.} Per-head attention   
  locality heatmap showing the coexistence of local and long-range heads at this stage. See Appendix                              
  Fig.~\ref{fig:attention_locality} for a line plot of attention distance.}                                                       
      \label{fig:attention_analysis}                                                                                              
  \end{figure}                                                                                                                    
                
\begin{table}[h]
  \caption{\textbf{Local attention suppression at the Physics Emergence Zone degrades performance.} Direction and intuitive physics degrade strongly under spatiotemporal local attention suppression on the Physics Emergence Zone, while ImageNet performance remains largely unchanged. For a full spatiotemporal sweep across a variety of $s$ and $t$, see Appendix Tab.~\ref{app:tab:attention-ablation}.}
  \label{tab:pez_knockout}
  \centering
  \vspace{-.25em}
  \begin{scriptsize}
    \begin{sc}
      \setlength{\tabcolsep}{4pt}
      \renewcommand{\arraystretch}{1.05}
      \begin{tabular}{@{}lccc@{}}
        \toprule
        Condition & Dir.\ ($R^2$) & IntPhys (\%) & INet (\%) \\
        \midrule
        Base                & 0.97 & 78.3 & 33.7 \\
        Spatial ($s{=}7$)   & 0.93 & 62.2 & 33.5 \\
        Temporal ($t{=}3$)  & 0.83 & \textbf{51.9} & 30.3 \\
        Combined ($s{=}3,t{=}1$)
                            & \textbf{0.14} & 61.7 & 33.1 \\
        \bottomrule
      \end{tabular}
    \end{sc}
  \end{scriptsize}
    \vspace{-.5em}                                            \end{table}

\begin{figure*}[t]                                                                                                              
      \centering                                                                                                                  
      \begin{subfigure}[b]{0.29\textwidth}                                                                                        
          \centering                                                                                                              
          \includegraphics[width=\textwidth]{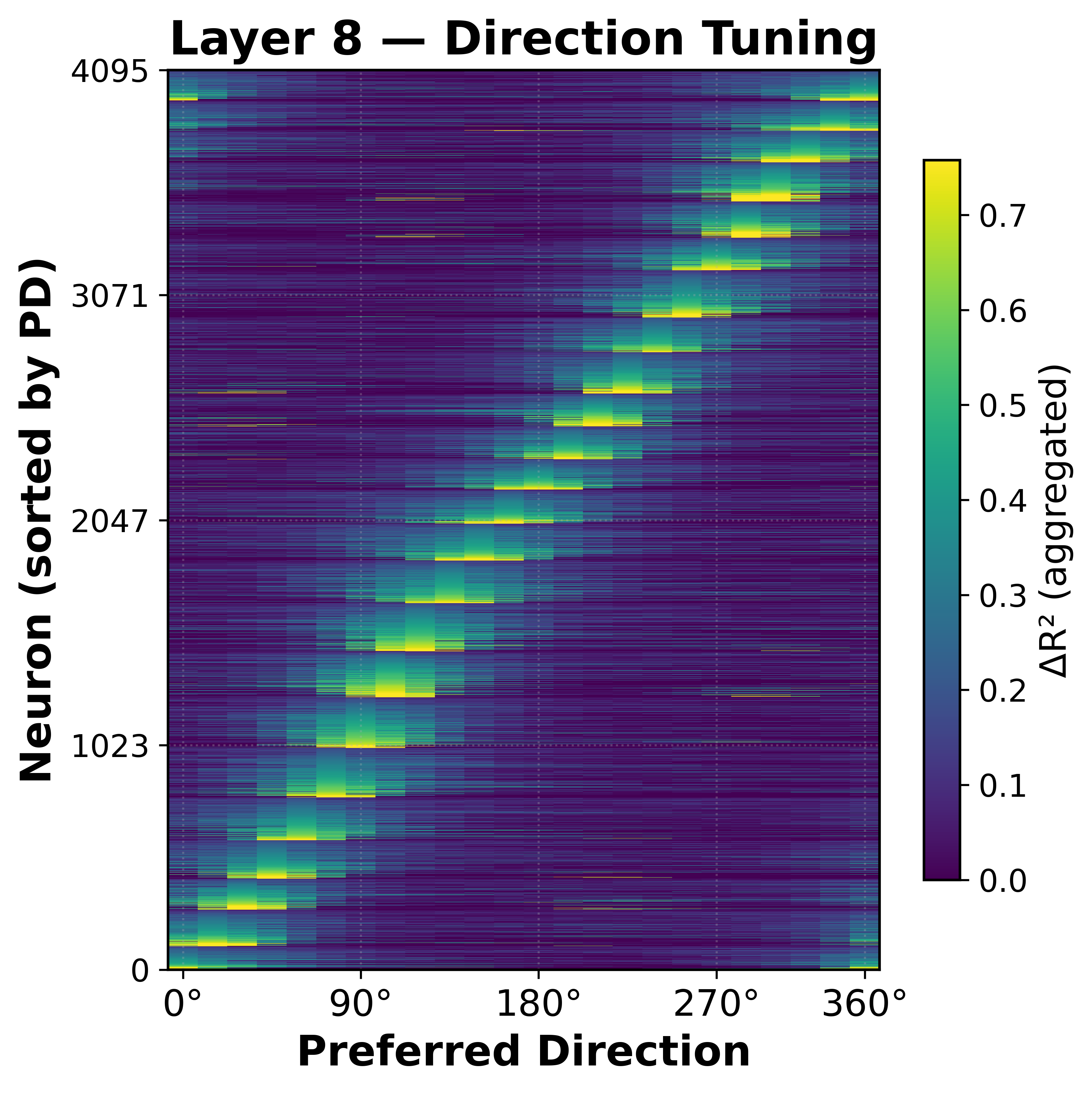}                                         
          \caption{}                                                                                                              
          \label{fig:unit_circle_heatmap}                                                                                         
      \end{subfigure}                                                                                                             
      \hfill                                                                                                                      
      \begin{subfigure}[b]{0.29\textwidth}                                                                                        
          \centering                                                                                                              
          \includegraphics[width=\textwidth]{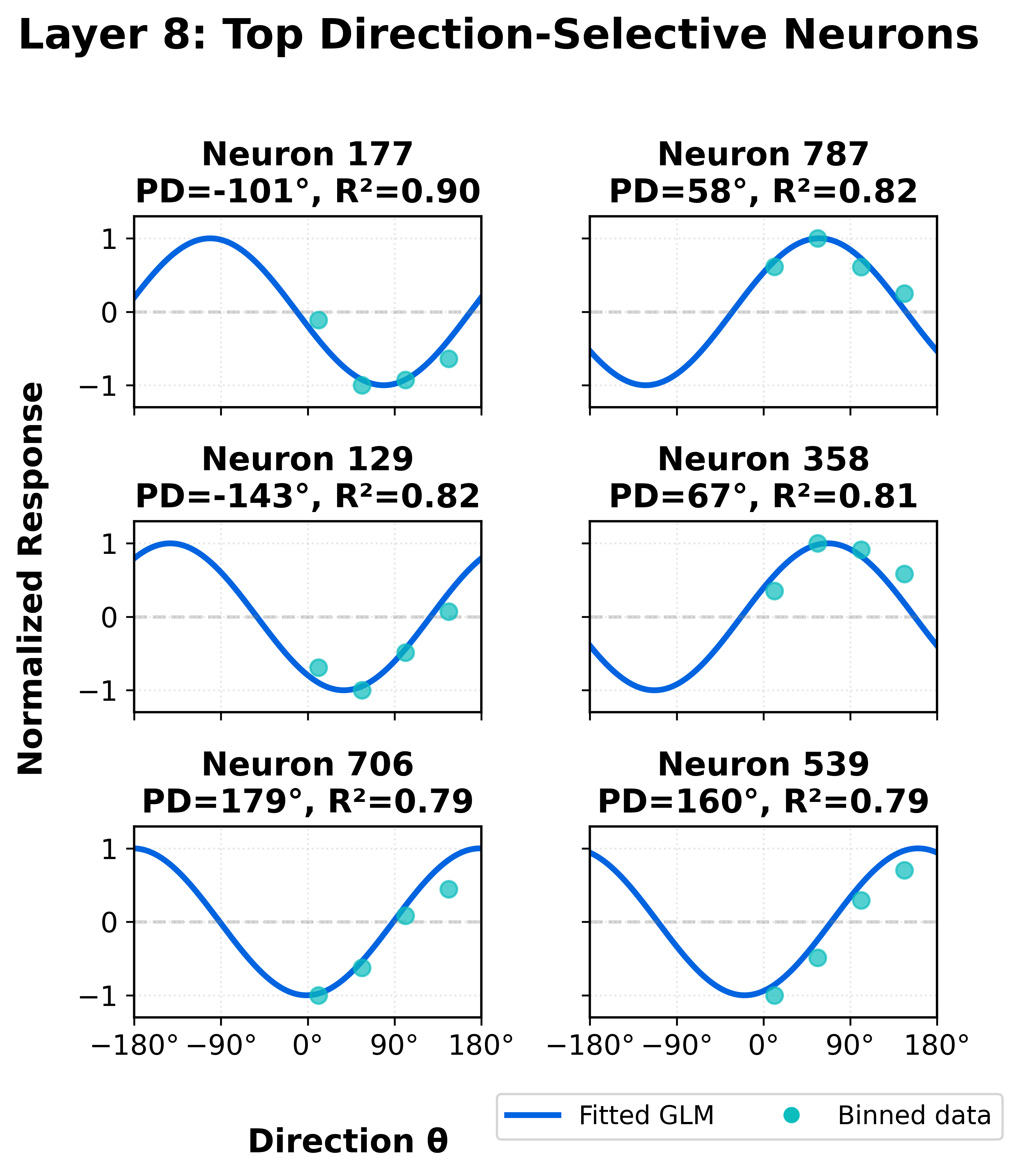}                                
          \caption{}                                                                                                              
          \label{fig:tuning_curves}                                                                                               
      \end{subfigure}                                                                                                             
      \hfill                                                                                                                      
      \begin{subfigure}[b]{0.33\textwidth}                                                                                        
          \centering                                                                                                              
          \includegraphics[width=\textwidth]{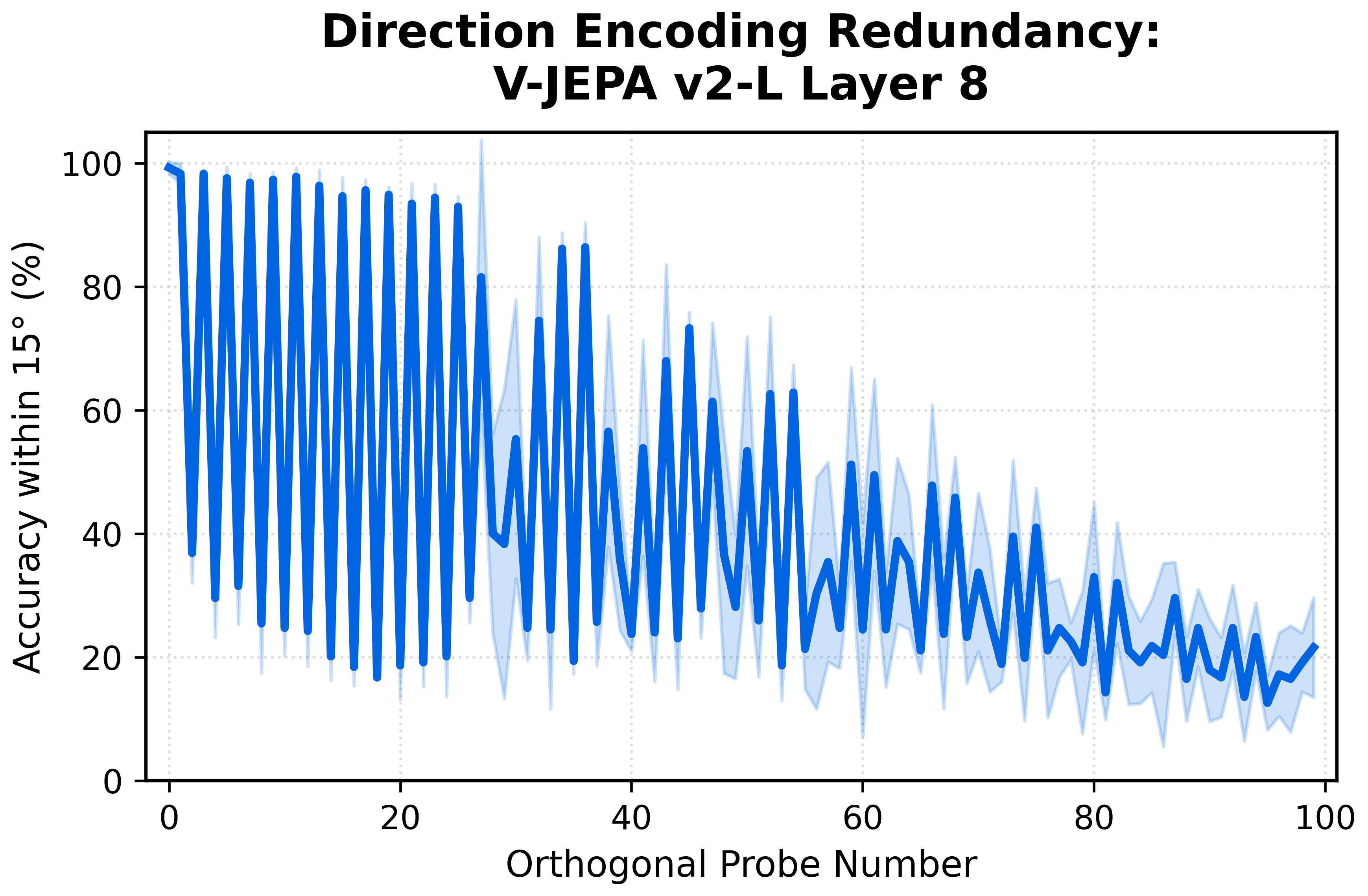}                         
          \caption{}                                                                                                              
          \label{fig:redundancy_sawtooth}                                                                                         
      \end{subfigure}                                                                                                             
      \caption{\textbf{Direction neurons form a ring-shaped population code with structured redundancy.} (a) At the one-third     
  emergence zone, direction-selective MLP units tile the full angular space and organize into a circular population code. (b)     
  Individual neurons exhibit smooth, sinusoidal tuning to motion direction. (c) Probe accuracy across successive                  
  orthogonalizations exhibits a sawtooth pattern, indicating structured redundancy consistent with paired (e.g., sine–cosine)     
  feature encodings.}                                                                                                             
      \label{fig:direction_unit_circle}                                                                                           
  \end{figure*}    
  
We next test whether local attention is functionally responsible for spatiotemporal processing by suppressing local attention exclusively in the Physics Emergence Zone (Appendix~\ref{app:appendix_attention_distance}). Our targeted suppression produces severe degradation in both direction decoding and possible-vs-impossible discrimination, while leaving the static task of ImageNet classification largely unaffected (Tab.~\ref{tab:pez_knockout}).

Our results identify a shared computational mechanism at the Physics Emergence Zone that supports the possible-vs-impossible physics and motion direction  tasks without requiring shared representational subspaces. Our analysis focuses on coarse spatiotemporal reasoning; other compositional physical computations, such as contact dynamics or force-based inference, may rely on additional mechanisms not examined here.

% ==================================================
\section{Steering the Direction Variable}
\label{sec:steering-motion-direction}

  % \begin{figure*}[h]                                                                                                              
  %     \centering                                                                                                                  
  %     \includegraphics[height=6cm]{paper/figures/unit_circle/unit_circle_heatmap.png}                                                   
  %     \hspace{1cm}                                                                                                                
  %     \includegraphics[height=6cm]{paper/figures/unit_circle/layer_8_neuron_tuning_curves.png}                                          
  %     \caption{Direction neurons form a ring-shaped population code. At the one-third emergence zone, direction-selective MLP     
  % units tile the full angular space and organize into a circular population code (left). Individual neurons exhibit smooth,       
  % sinusoidal tuning to motion direction (right).}                                                                                 
  %     \label{fig:direction_unit_circle}                                                                                           
  % \end{figure*}   

One gold standard in interpretability is the ability to steer representations in latent space to change the model's prediction, establishing the feature's causal impact on the output \citep{turner2024steeringlanguagemodelsactivation, panickssery2024steeringllama2contrastive, zou2025representationengineeringtopdownapproach}. In the following section, we attempt to steer motion direction to change the decoding. We find a rich and counterintuitive set of behaviors for the representation of direction: circular population geometry, "sawtooth" sin-cosine encoding, and successful steering only along manipulating dozens of orthogonal probe dimensions.

% \subsection{Direction Representations Become Patch-Accessible at the Physics Emergence Zone}
% \label{sec:direction-and-patches}

% To characterize what changes at the Physics Emergence Zone, we analyze the spatiotemporal organization of direction representations using patch-level probes. In early layers, direction information is fragmented across patches: individual patches do not support reliable decoding, although mean-pooled probes recover weak signals by aggregating across space and time (Appendix Fig.~\ref{fig:per_patch_velocity}a). At the Physics Emergence Zone, direction becomes redundantly accessible across patches, producing a sharp increase in per-patch decoding performance and spatial generalization (Appendix Fig.~\ref{fig:per_patch_velocity}a,b). After this transition, direction can be decoded reliably from individual patches, even when they do not contain the moving object.

\subsection{Direction neurons form a ring-shaped population code at the Physics Emergence Zone}
\label{sec:direction-ring}

We find that MLP layers (fc1/fc2) at the end of the Physics Emergence Zone selectively encode motion direction (Fig.~\ref{fig:unit_circle_heatmap}), with most units strongly direction-tuned (high GLM $R^2$), and preferred directions tiling $360^\circ$ (Fig.~\ref{fig:tuning_curves}). At the population level, direction-selective features organize into a unit-circle geometry that is absent in early layers and emerges sharply at the transition (Appendix Fig.~\ref{fig:layer-0-layer-8}), consistent with a circular population code in motion neuroscience \citep{jazayeri2006optimal}. We do not observe an analogous circular organization for speed, indicating a qualitative difference between direction and speed (Appendix Fig.~\ref{app:fig:speed_unit_circle}). For our full methodology, see Appendix~\ref{app:direction_tuning}.

% \begin{figure}[h]                                                                                                              
%       \centering                                                                                                                  
%       \includegraphics[width=\columnwidth, height=5cm, keepaspectratio]{paper/figures/unit_circle/unit_circle_heatmap.png}                                                   
             
%       \vspace{0.3em}                                                                                                               
%          \includegraphics[width=\columnwidth, height=5cm, keepaspectratio]{paper/figures/unit_circle/layer_8_neuron_tuning_curves.png}                                   
%       \caption{Direction neurons form a ring-shaped population code. At the one-third emergence zone, direction-selective MLP     
%   units tile the full angular space and organize into a circular population code (top). Individual neurons exhibit smooth,        
%   sinusoidal tuning to motion direction (bottom).}                                                                                
%       \label{fig:direction_unit_circle}                                                                                           
%   \end{figure}  

However, manipulating only the unit-circle subspace does not effectively steer direction, suggesting that direction is embedded in a higher-dimensional representation beyond a single unit  circle. We test this hypothesis in the next section.

% The neuroscience analogy holds with the globalization of direction information. To characterize what changes at the Physics Emergence Zone, we analyze the spatiotemporal organization of direction representations using patch-level probes. In early layers, direction information is fragmented across patches: individual patches do not support reliable decoding, although mean-pooled probes recover weak signals by aggregating across space and time (Appendix Fig.~\ref{fig:per_patch_velocity}a). At the Physics Emergence Zone, direction becomes redundantly accessible across patches, producing a sharp increase in per-patch decoding performance and spatial generalization (Appendix Fig.~\ref{fig:per_patch_velocity}a,b). After this transition, direction can be decoded reliably from individual patches, even when they do not contain the moving object. 

\subsection{Physics-related variables require many tens of directions to steer}
\label{sec:high-feature-dimensionality}

To estimate the dimensionality of motion direction, we iteratively train linear probes, orthogonalize each probe direction, and retrain on the residual representation until performance reaches chance, a method closely related to prior work on iterative nullspace projection and amnesic probing \citep{ravfogel-etal-2020-null}. We describe our method in detail in Appendix~\ref{app:orthogonal_probes_method}.

% \begin{figure}[h]
%       \centering
%       \begin{subfigure}[b]{0.48\textwidth}
%           \centering
%           \includegraphics[width=\textwidth]{paper/figures/feature_dimensionality/fig_orthogonal_probe_counts_by_task.png}
%   \caption{Estimated feature dimensionality of decoded variables across tasks, measured by the number of orthogonal linear probes trainable before performance approaches   
%   chance (Direction: $R² < 0.3$; Speed: $R² < 0.1$; IntPhys: accuracy $< 55\%$). Physical variables require tens of independent features.}                                        
%           \label{fig:feature_dimensionality_count}
%       \end{subfigure}
%       \hfill
%       \begin{subfigure}[b]{0.48\textwidth}
%           \centering
%           \includegraphics[width=\textwidth]{paper/figures/feature_dimensionality/fig_redundancy_acc.png}
%           \caption{Probe accuracy across successive orthogonalizations exhibits a sawtooth pattern, indicating structured redundancy consistent with paired (e.g., sine–cosine) feature encodings.}
%           \label{fig:sin-cos-sawtooth}
%       \end{subfigure}
%         \caption{High-dimensional encoding of physical variables. Direction and intuitive physics are represented across dozens of nearly orthogonal features rather than a small set of units (a), with structured redundancy consistent with paired feature encodings (b).}
%   \label{fig:feature_dimensionality}
% \end{figure}

We find that speed, direction, and possible-vs-impossible representations are encoded in high-dimensional subspaces. Possible-impossible discrimination requires approximately 20 independent features at the Physics Emergence Zone, while direction decoding requires roughly 40--50 features, increasing to up to 80 near the output layers (Appendix Fig.~\ref{fig:feature_dimensionality_count}). Interestingly, probe performance exhibits a characteristic sawtooth pattern across successive orthogonalizations (Fig.~\ref{fig:redundancy_sawtooth}), consistent with direction being encoded via approximately sinusoidal feature pairs (e.g., sine--cosine components) -- a pattern that we see uniquely in motion direction and not speed (Appendix Fig.~\ref{app:fig:speed-sawtooth}).

We next examine whether the high-dimensional direction representation can be causally controlled along its many feature dimensions. In line with the previous 
  section's results about manipulating the unit circle in the MLP layers, steering along a single feature direction or probe axis produces little to no change  
  in decoded motion direction. In contrast, coordinated interventions across increasing numbers of orthogonal probe directions yield smooth and monotonic       
  reductions in mean angular error (Appendix, Fig.~\ref{fig:steering_generalization}). When steering across all probe directions at layer 8, we achieve           
  $<0.5^\circ$ error to target angles, compared to $>80^\circ$ for single-probe interventions (Appendix~\ref{app:steering}). Effective steering therefore
   requires manipulating a large fraction of the representational subspace, rather than modifying any individual feature. 

Surprisingly, unlike language models, where semantic concepts often align with single or low-rank directions, even complex behaviors such as refusal can often be controlled via a single activation direction or a small set of vectors \citep{turner2024steeringlanguagemodelsactivation, panickssery2024steeringllama2contrastive, zou2025representationengineeringtopdownapproach, arditi2024refusallanguagemodelsmediated}, motion direction in video encoders requires coordinated high-dimensional steering.

% ==================================================

\section{Discussion}
\label{sec:discussion}

Our goal in this work was to characterize the \emph{representational form} through which physical information is made available internally.

\subsection{Intuitive physics in cognitive science}
\label{sec:intuitive-physics-discussion}

A central debate in cognitive science concerns whether intuitive physics relies on compact, reusable latent state variables—akin to a physics engine—or on distributed, task-specific computations \citep{battaglia2013simulation,ullman2017mind,davis2017commonsense}. Our results support the latter at the representational level. Despite strong physical behavior, we find no evidence for shared, low-dimensional latent variables: motion direction and possible–impossible judgments occupy nearly orthogonal subspaces, and direction itself requires high-dimensional coordinated steering (Tab.~\ref{tab:mechanistic_predictions}). These findings are difficult to reconcile with physics-engine-style representations and instead support distributed, task-specific representations built atop shared spatiotemporal computation.

\subsection{Connections to neuroscience}
\label{sec:connections-to-neuroscience}

The organization of motion direction we observe closely parallels biological vision. In primate cortex, direction-selective neurons tile angular space and direction is represented as a circular population code rather than an explicit latent variable \citep{albright1984direction,jazayeri2006optimal}. Similarly, direction in video world models emerges as a distributed circular geometry. Moreover, the early availability of speed and later emergence of direction mirror the motion-processing hierarchy in cortex, where direction selectivity arises through higher-order pooling \citep{born2005structure,pasternak2020linking}. Together, these parallels suggest that video world models may provide an ideal complementary substrate for studying findings in neuroscience.

\subsection{Applications to physics simulators}

 Recent work on physics steering shows that a transformer trained on PDE simulations (not video) can admit low-dimensional activation interventions corresponding to interpretable physical phenomena \citep{mccabe2025walruscrossdomainfoundationmodel}. In contrast, our results show that our video world models encode motion variables in distributed, high-dimensional geometries, suggesting that whether learned models expose compact, interpretable “state variables” is not guaranteed, but depends critically on training domain and objective.

% ==================================================
\section{Limitations}

Our analysis is limited to encoder-based video transformers trained with masked objectives; autoregressive or diffusion video models may exhibit different representational structure. We focus on possible--impossible discrimination and controlled motion variables, which probe core spatiotemporal sensitivity but do not isolate richer physical computations such as contact dynamics, force inference, or long-horizon interaction. Our methods characterize representational accessibility and coarse causal influence rather than a complete circuit-level mechanism, and our synthetic toy-ball dataset may not reflect how physical structure is represented in natural video.

\section{Conclusion}

We map where and how physics-relevant information appears inside large-scale video encoders. Across V-JEPA 2 and VideoMAE-v2 G, both possible--impossible discrimination and motion direction emerge at a sharp mid-depth transition—the \emph{Physics Emergence Zone}—after which physics signals peak and then weaken toward the output layers. Motion decomposition reveals a clear asymmetry: scalar magnitudes are available early, while direction becomes linearly accessible only at this transition. Although these abilities co-emerge, direction and possible--impossible judgments occupy nearly orthogonal representational subspaces, while both depend causally on localized spatiotemporal attention. Direction itself is encoded as a high-dimensional, unit circle population code that resists low-dimensional steering. These findings argue against compact, reusable latent physics state and instead support a distributed, task-specific representational regime, in favor of heuristic-based accounts of intuitive physics proposed in cognitive science.

% % ==================================================
% \section{Conclusion}

% We show that physical reasoning in video world models emerges from an intermediate computational regime and is supported by distributed population-based representations rather than compact reusable state variables.

% Acknowledgements only in camera-ready version
% \section*{Acknowledgements}
% Your acknowledgements.

\section*{Impact Statement}

This paper presents work whose goal is to advance the field of Machine Learning. There are many potential societal consequences of our work, none of which we feel must be specifically highlighted here.

\bibliography{paper}
\bibliographystyle{icml2026}

% Appendix
\clearpage
\onecolumn
\appendix

\section{Experimental Details}

\subsection{Dataset Details}

\subsubsection{Intuitive Physics}
\label{app:intuitive_physics_dataset_details}
\begin{figure*}[h]
  \centering
  \includegraphics[width=.77\textwidth]{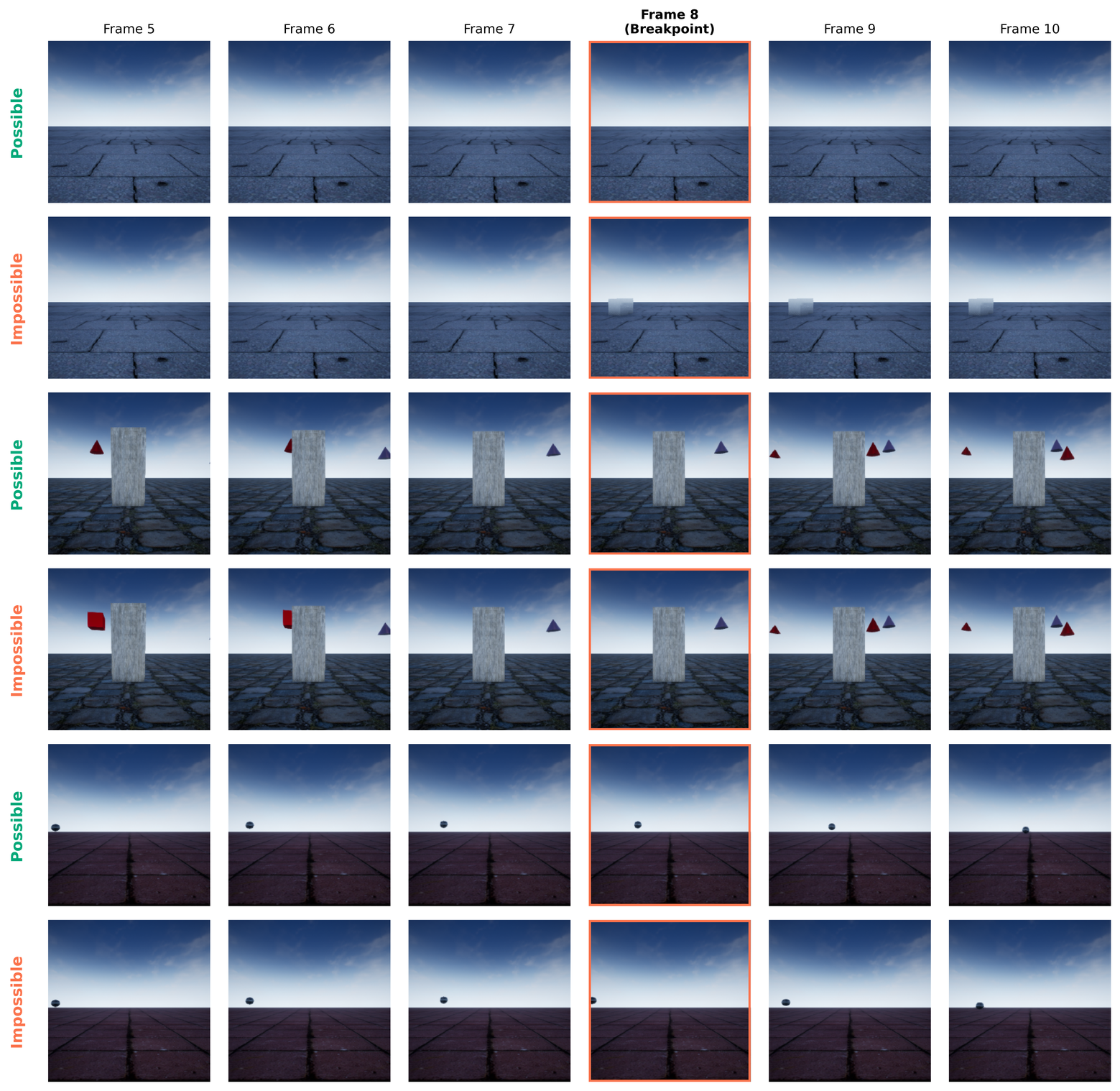}
  \caption{A possible and impossible example from each fold of IntPhys.}
  \label{app:fig:intuitive_phyiscs_dataset}
\end{figure*}

\subsubsection{Synthetic Ball Dataset}
\label{app:dataset_details:toy_ball}

We generate two controlled synthetic motion datasets using the Kubric physics simulator \cite{greff2022kubric} to study representations of constant-velocity and uniformly accelerated motion. All videos depict a single sphere moving along straight-line trajectories with known ground-truth dynamics.

\paragraph{Velocity dataset.}

The velocity dataset contains 392 videos (8 directions $\times$ 7 speeds $\times$ 7 start positions), each 16 frames long at 24 fps (0.67 s), rendered at $256 \times 256$ resolution. Motion directions are $\theta \in \{0^\circ,45^\circ,\dots,315^\circ\}$, with speed magnitudes $v \in \{1,\dots,7\}$ m/s. Start positions are sampled uniformly from $[-2,2]^2$ m for each $(\theta,v)$ pair.

Scenes are generated using Kubric \cite{greff2022kubric}, with physics simulated in PyBullet 3.2.5 and rendering performed in Blender 2.93. Each scene contains a sphere (radius 0.3 m, mass 1.0 kg), a static $8 \times 8$ m floor plane, a fixed overhead perspective camera at (0,0,10) m, and a single directional light.

All friction terms (lateral, rolling, spinning) and restitution are set to zero. The sphere is initialized with nonzero velocity $v$ in direction $\theta$, and no external forces are applied, yielding uniform linear motion throughout the sequence ($\Delta v = 0$).

\paragraph{Acceleration dataset.}

The acceleration dataset consists of 280 videos (8 directions $\times$ 5 accelerations $\times$ 7 start positions) with identical temporal and spatial resolution. Motion directions follow the same $\theta$ set as above, with acceleration magnitudes $a \in \{2,4,6,8,10\}$ m/s$^2$. Start positions are sampled uniformly from $[-2,2]^2$ m for each $(\theta,a)$ pair.

The sphere is initialized at rest and subjected to a constant external force $\mathbf{F}=m\mathbf{a}$ applied at each timestep in direction $\theta$. Physics is simulated at 240 Hz (10 substeps per rendered frame). Under frictionless conditions, measured accelerations match target values within $10^{-4}$ m/s$^2$, verified via finite differences of recorded velocities. Rendering, scene configuration, and annotations are identical to the velocity dataset. 

% \subsection{Dataset Comparison}

% \begin{table}[h]
% \centering
% \renewcommand{\arraystretch}{0.9}
% \begin{tabular}{lll}
% \toprule
% \textbf{Property} & \textbf{Velocity} & \textbf{Acceleration} \\
% \midrule
% Total videos & 392 & 280 \\
% Initial velocity & $1$--$7$ m/s & 0 m/s \\
% Applied force & None & Constant $F=ma$ \\
% Motion type & Uniform linear & Uniform acceleration \\
% Velocity change & $\Delta v=0$ & $\Delta v=a t$ \\
% Distance traveled & $v t$ & $\tfrac{1}{2} a t^2$ \\
% \bottomrule
% \end{tabular}
% \caption{Comparison of synthetic motion datasets.}
% \end{table}

\section{Probe Training Details}
\label{app:probe_training_details}

For all experiments, we trained linear probes of the form
$f(\bm{h}_\ell) = \bm{W}\bm{h}_\ell + \bm{b}$
on spatiotemporally pooled activations from each layer
$\ell \in \{0, \ldots, n-1\}$ of the frozen $n$-layer encoder.
We performed a hyperparameter sweep over 20 configurations,
with learning rates
$\{10^{-4}, 3 \times 10^{-4}, 10^{-3}, 3 \times 10^{-3}, 5 \times 10^{-3}\}$
and weight decay
$\{0.01, 0.1, 0.4, 0.8\}$,
selecting the best model based on validation performance.
We used 5-fold grouped cross-validation and report results as
mean $\pm$ standard deviation across folds.

\FloatBarrier

\section{Additional Experiments}

\subsection{Possible-vs-impossible physics task}
\label{app:additional_experiments:possible-vs-impossible}

\subsubsection{Full results for linear probe}

  \begin{figure}[h]
      \centering
      \includegraphics[width=.9\columnwidth]{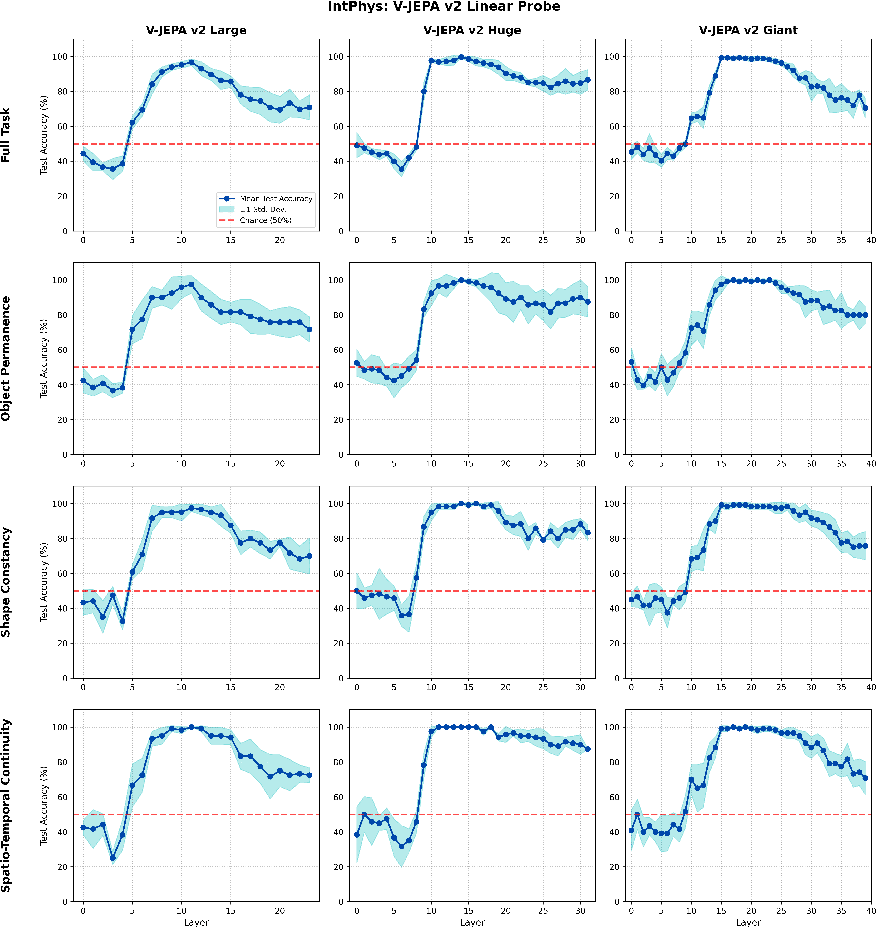}
      \caption{Full results for the linear probe on all sizes of V-JEPA 2.}
      \label{app:fig:intphys_all_jepav2_linear}
  \end{figure}

    \begin{figure}[h]
      \centering
      \includegraphics[width=\columnwidth]{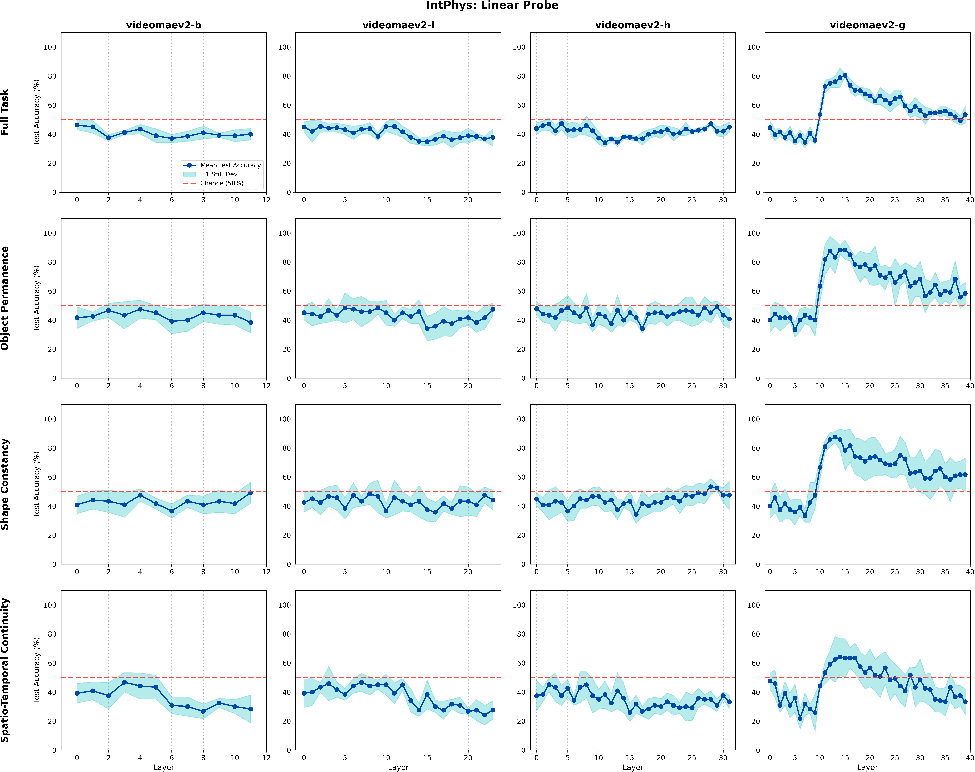}
      \caption{Full results for the linear probe on all sizes of VideoMAE-v2.}
      \label{app:fig:intphys_all_videomaev2_linear}
  \end{figure}

\FloatBarrier
\subsubsection{Full results for attentive-MLP probe}

  \begin{figure}[h]
      \centering
      \includegraphics[width=.8\columnwidth]{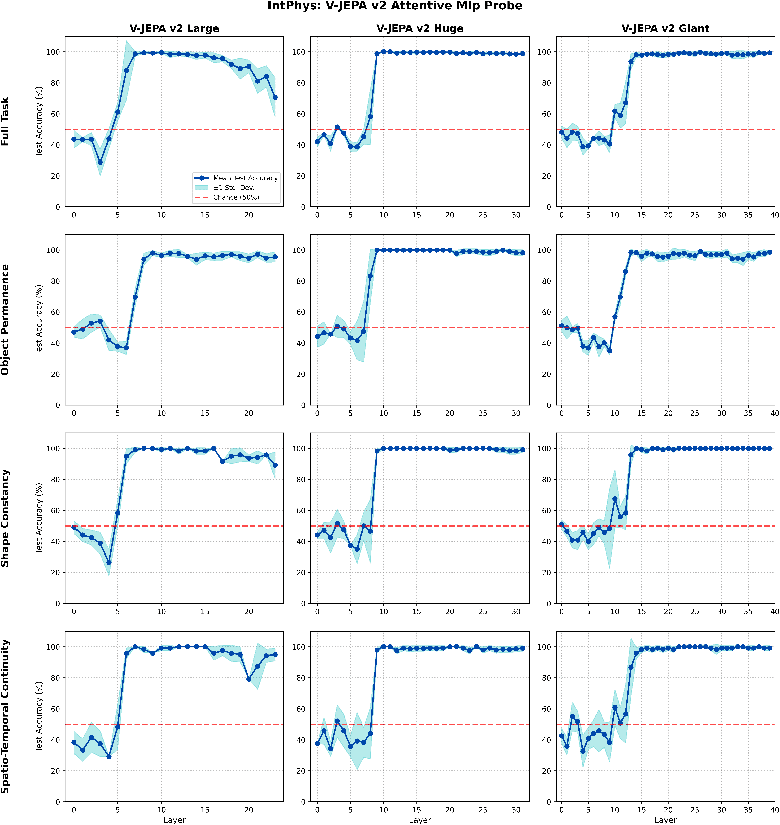}
      \caption{Full results for the attentive-mlp probe on all sizes of V-JEPA 2.}
      \label{app:fig:intphys_all_jepav2_attentive_mlp}
  \end{figure}

    \begin{figure}[h]
      \centering
      \includegraphics[width=\columnwidth]{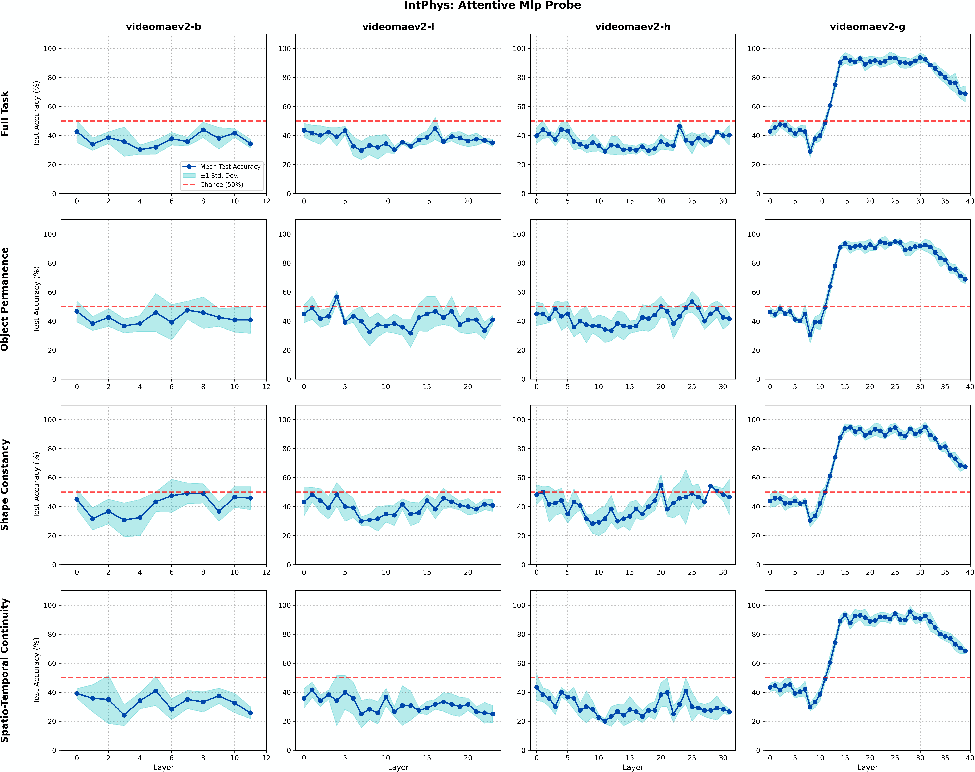}
      \caption{Full results for attentive-mlp probe on all sizes of VideoMAE-v2.}
      \label{app:fig:intphys_all_videomaev2_attentive_mlp}
  \end{figure}

\FloatBarrier
\subsubsection{Consistency across possible-vs-impossible physics sub-tasks}
\label{app:consistency-across-intphys-subtasks}

\begin{figure}[H]
   \centering
   \includegraphics[width=0.66\textwidth]{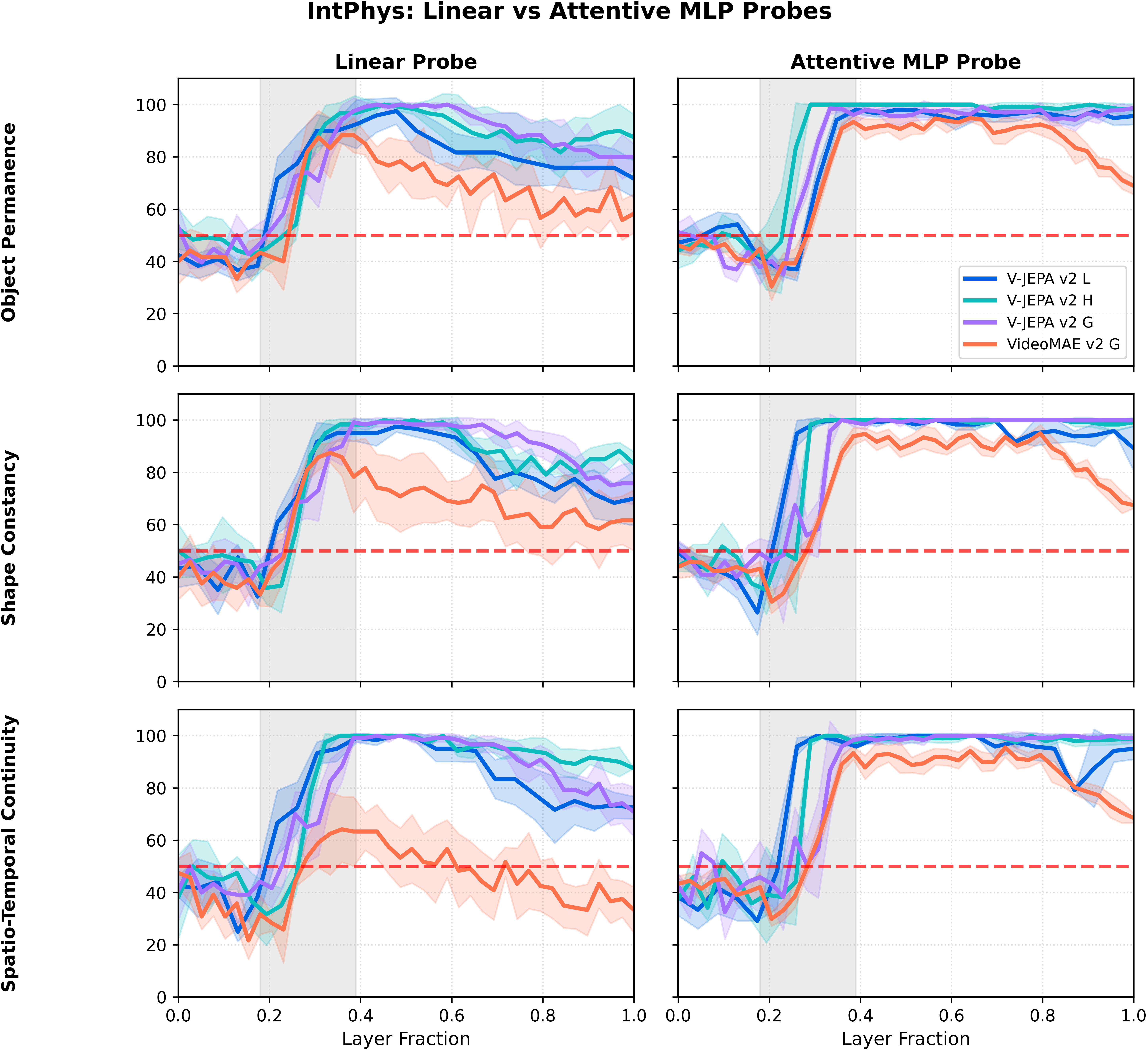}
   \caption{Intuitive physics results broken down by subtasks of object permanence, shape constancy, and spatiotemporal continuity.}
   \label{fig:intphys_data_subtask}
\end{figure} 

In Section~\ref{sec:intuitive-physics-emergence}, we show a distinct emergence pattern one-third through the network for the possible vs. impossible physics task. The IntPhys~\cite{riochet2021intphys} dataset is further divided into three sub-tasks capturing object permanence, shape constancy, and spatiotemporal continuity. Therefore, a natural next question is whether each sub-task has a unique emergence signature.

Probe performance by subtask reveals the same one-third emergence pattern across all three principles: object permanence, shape
constancy, and spatiotemporal continuity (Figure~\ref{fig:intphys_data_subtask}). This universality suggests that the models may have the same underlying processing for possible vs. impossible tasks.

\subsubsection{The middle layer possible-vs-impossible physics representations generalize to better performance on a downstream intuitive physics task}
\label{app:strongest-representations-downstream-task}

\begin{figure}[h]
\centering
    \includegraphics[width=0.4\textwidth]{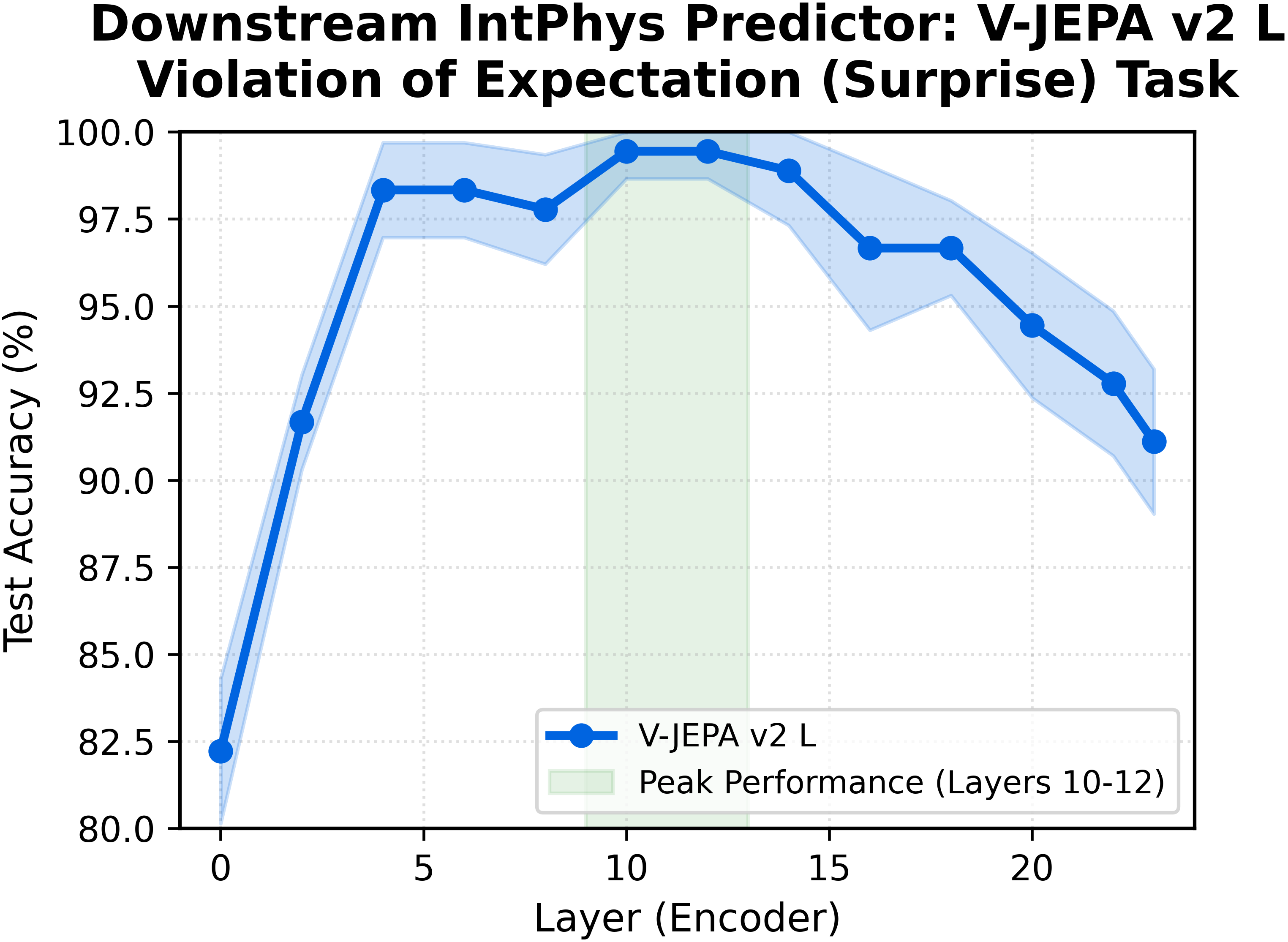}
    \caption{Downstream performance of the V-JEPA 2 Large predictor on the violation-of-expectation intuitive physics set-up. Representations from the middle layers of the encoder provide the best performance compared to final layers.}
    \label{fig:intphys_predictor_performance}
\end{figure}

In Section~\ref{sec:polar-results}, we show that perhaps counterintuitively strongest representations for the possible vs. impossible intuitive physics task are at the center of the network across model sizes and architectures. Our findings echo past results in which intermediate instead of final representations were seen to be useful on certain tasks for contrastively vision-language models \cite{bolya2025perceptionencoderbestvisual}. Similarly, using intermediate instead of end representations may be useful in tasks like using V-JEPA-2 representations to improve the physics plausibility of video generative models \cite{yuan2025improvingphysicsvideogeneration}.

To confirm this hypothesis, we trained from scratch a V-JEPA 2 large predictor on the representations of every layer on the violation-of-expectation framework taken from~\cite{garrido2025intuitivephysicsunderstandingemerges}. This downstream task measures the plausibility of the next-frame, and it is a more real-world task than our previous set-up.

We confirm that the middle one third of the network yields the best predictivity on the task (Appendix Figure \ref{fig:intphys_predictor_performance}). This suggests that for future tasks involving training on physical representations, the center of the network may be more optimal than the end of the network, where performance degrades.

Interestingly, we also notice that even early layers that give poor performance on only the encoder suddenly perform well (Layer 0). This is likely because the predictor itself is learning from the representation.

Future interpretability work will need to investigate why this is the case; we hypothesize it may be related to feature object binding, in which velocity information is most "bound" to the corresponding object. At the end of the network, information is catered to the optimization objective of predicting the next frame in latent space, not necessarily to preserving object-level information.

\subsection{Task-Specificity of Physics Emergence Zone}

\begin{figure}[h]
   \centering
   \includegraphics[width=.66\columnwidth]{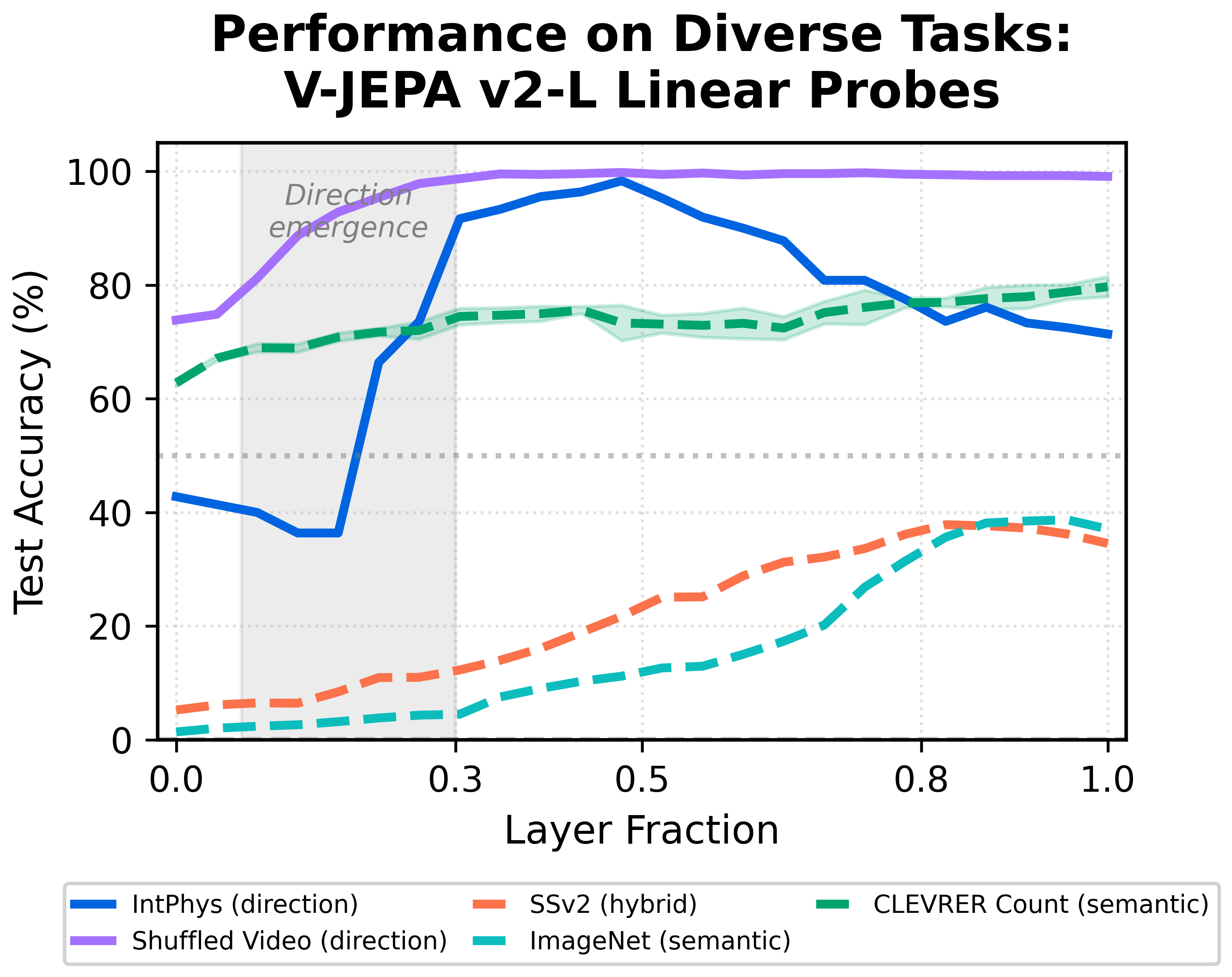}
   \caption{Layer-wise probe performance on control tasks. Object counting, image classification, and standard video classification do not exhibit the one-third emergence signature, indicating it is not a generic depth effect. Only shuffled video detection shows a similar pattern, consistent with relying on temporal order.}
   \label{fig:diverse_tasks}
\end{figure} 

\subsection{Direction validation on multi-object scenes}
\label{app:additional_experiments:clevrer}

\begin{figure}[h]                                                                                                               
      \centering                                                                                                                  
      \begin{subfigure}[b]{0.48\textwidth}                                                                                        
          \centering                                                                                                              
          \includegraphics[width=\textwidth]{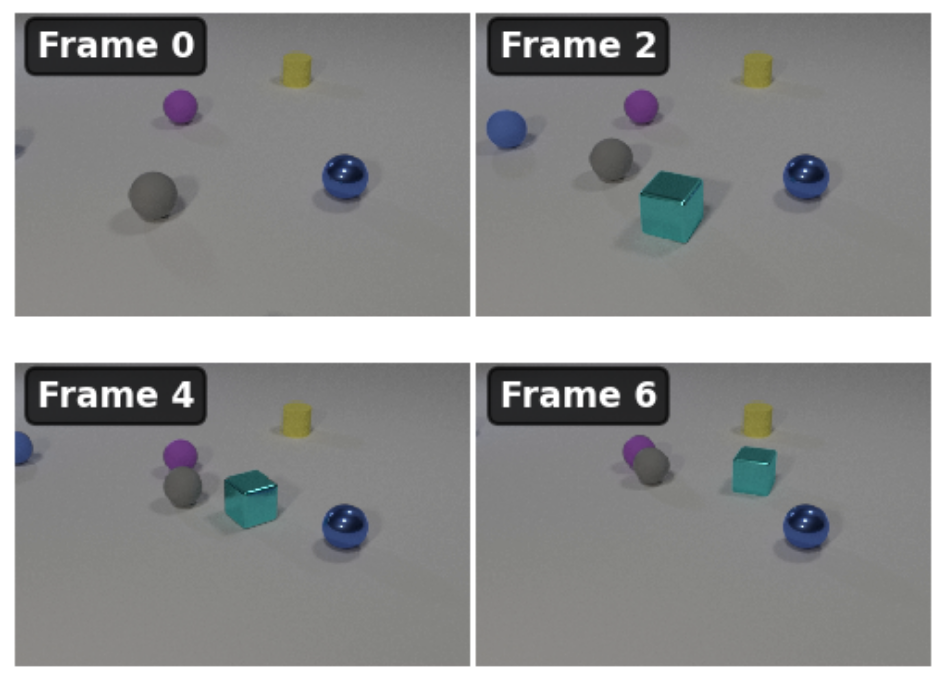}                                            
          \caption{}                                                                                                              
          \label{fig:clevrer_frames}                                                                                              
      \end{subfigure}                                                                                                             
      \hfill                                                                                                                      
      \begin{subfigure}[b]{0.48\textwidth}                                                                                        
          \centering                                                                                                              
          \includegraphics[width=\textwidth]{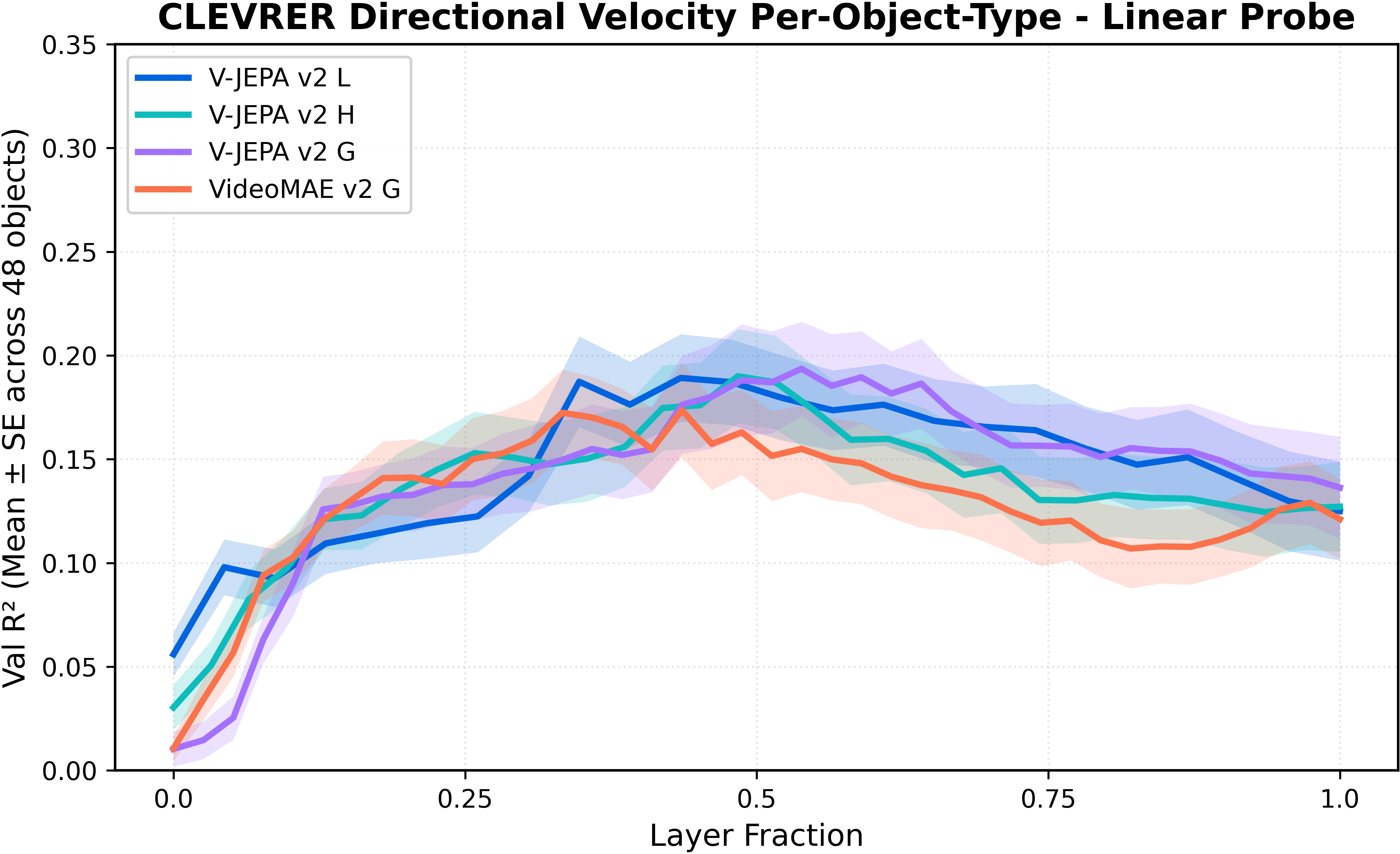}                                                      
          \caption{}                                                                                                              
          \label{fig:clevrer_r2}                                                                                                  
      \end{subfigure}                                                                                                             
      \caption{CLEVRER dataset: (a) Sample video frames. The first six frames of the CLEVRER video dataset, which shows objects of
   various colors, shapes, and textures moving across the screen. We train per-object probes on each of the objects to detect its 
  direction. (b) Per-object direction results.}                                                                                   
      \label{fig:clevrer_per_object_direction}                                                                                    
  \end{figure}

In Section~\ref{sec:polar-results}, we validated the emergence of direction for multi-object scenes like the CLEVRER dataset to discover that the Physics Emergence Zone signature was consistent across model architectures (Fig. \ref{fig:clevrer_per_object_direction} ) \cite{yi2020clevrercollisioneventsvideo}. This Appendix section contains additional results, including a per-object breakdown for accuracy.

\begin{figure}[h]
      \centering
      \includegraphics[width=.88\columnwidth]{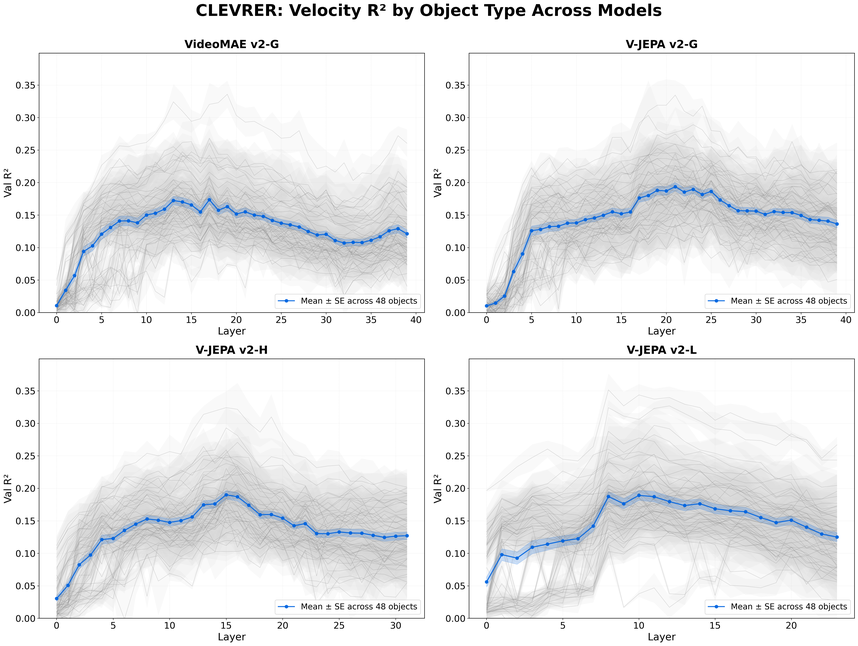}
      \caption{Full results for linear probes for all 48 objects of CLEVRER. Each gray line is an object and the blue is the average across all objects. Error bars are from k-fold cross validation.}
      \label{app:fig:clevrer_per_object_results_full_results}
\end{figure}

\begin{figure}[h]
      \centering
      \includegraphics[width=.88\columnwidth]{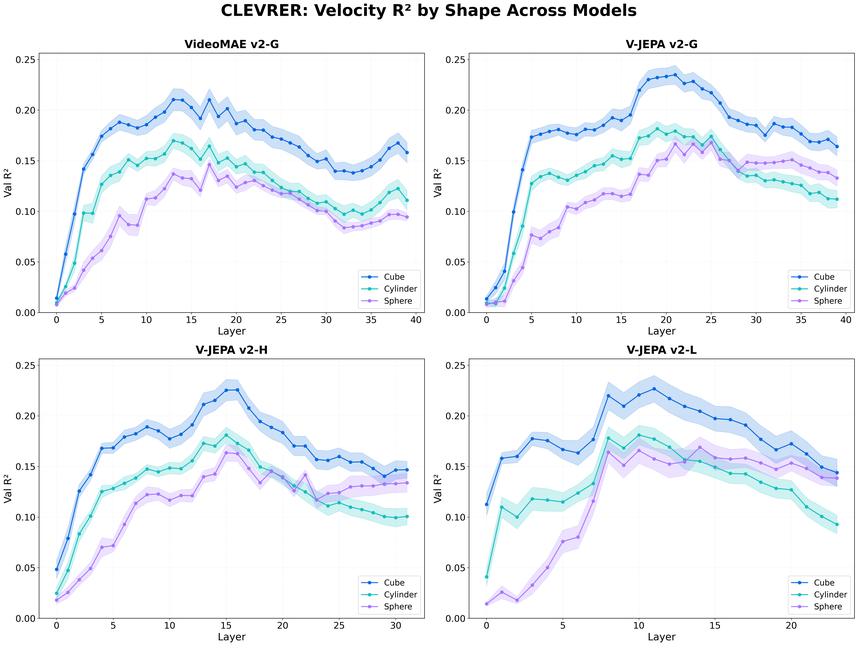}
      \caption{Results for CLEVRER velocity per-object linear probe R² grouped by cube, cylinder, and sphere shape across all model types.}
      \label{app:fig:clevrer_per_object_results_shape}
\end{figure}

\begin{figure}[h]
      \centering
      \includegraphics[width=.88\columnwidth]{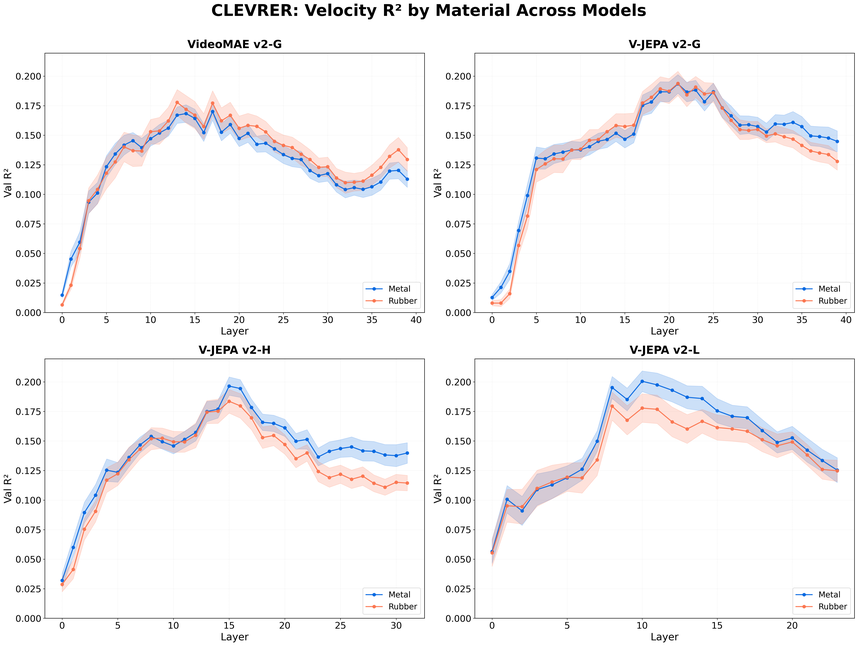}
      \caption{Results for CLEVRER velocity per-object linear probe R² grouped by metal and rubber material across all model types.}
      \label{app:fig:clevrer_per_object_results_material}
\end{figure}

  \begin{figure}[h]
      \centering
      \begin{subfigure}[b]{0.49\textwidth}
          \centering

  \includegraphics[width=\textwidth]{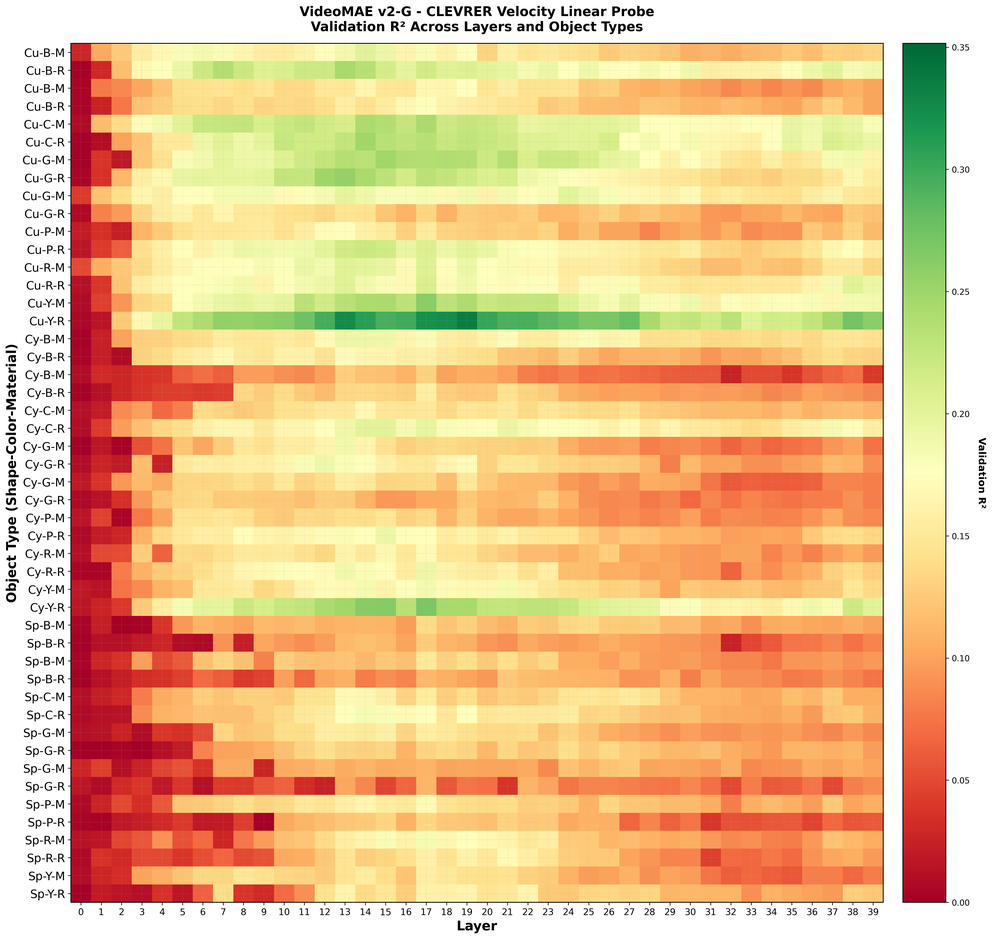}
          \caption{VideoMAE-v2-G}
          \label{fig:heatmap_videomaev2g}
      \end{subfigure}
      \hfill
      \begin{subfigure}[b]{0.49\textwidth}
          \centering

  \includegraphics[width=\textwidth]{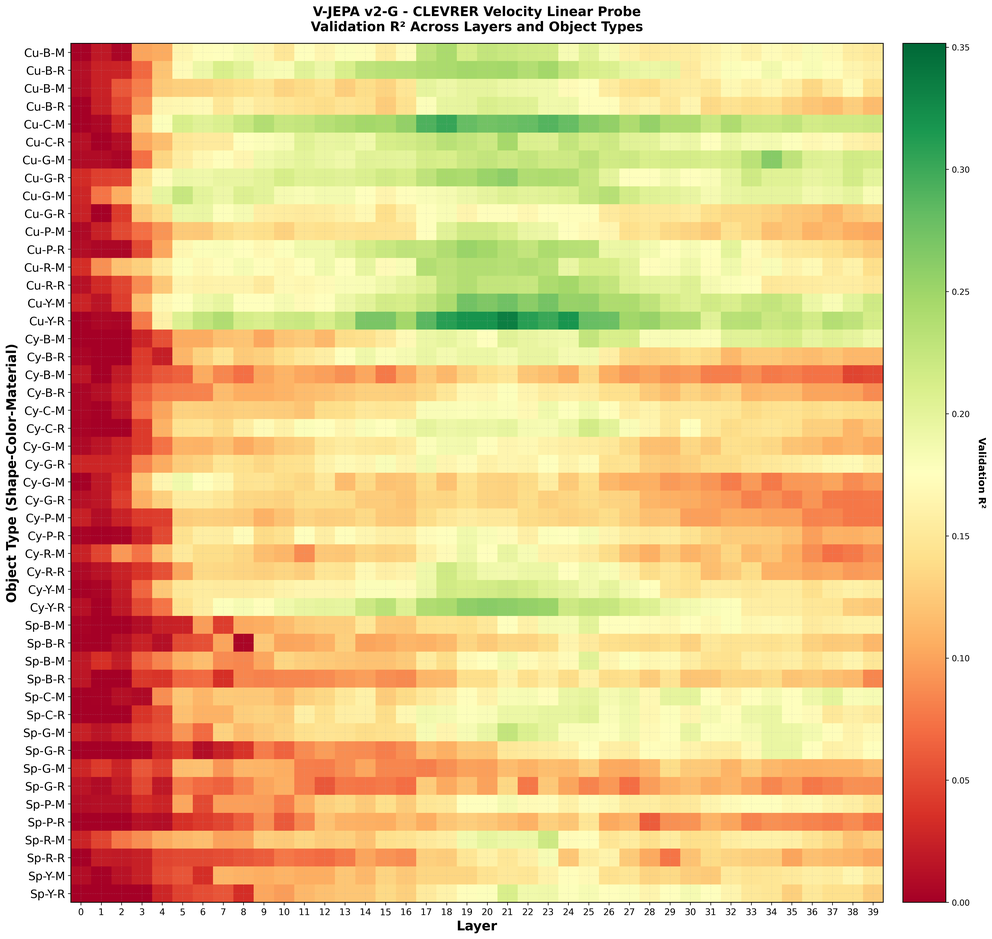}
          \caption{V-JEPA 2-G}
          \label{fig:heatmap_vjepav2g}
      \end{subfigure}

      \vspace{0.3cm}

      \begin{subfigure}[b]{0.49\textwidth}
          \centering

  \includegraphics[width=\textwidth]{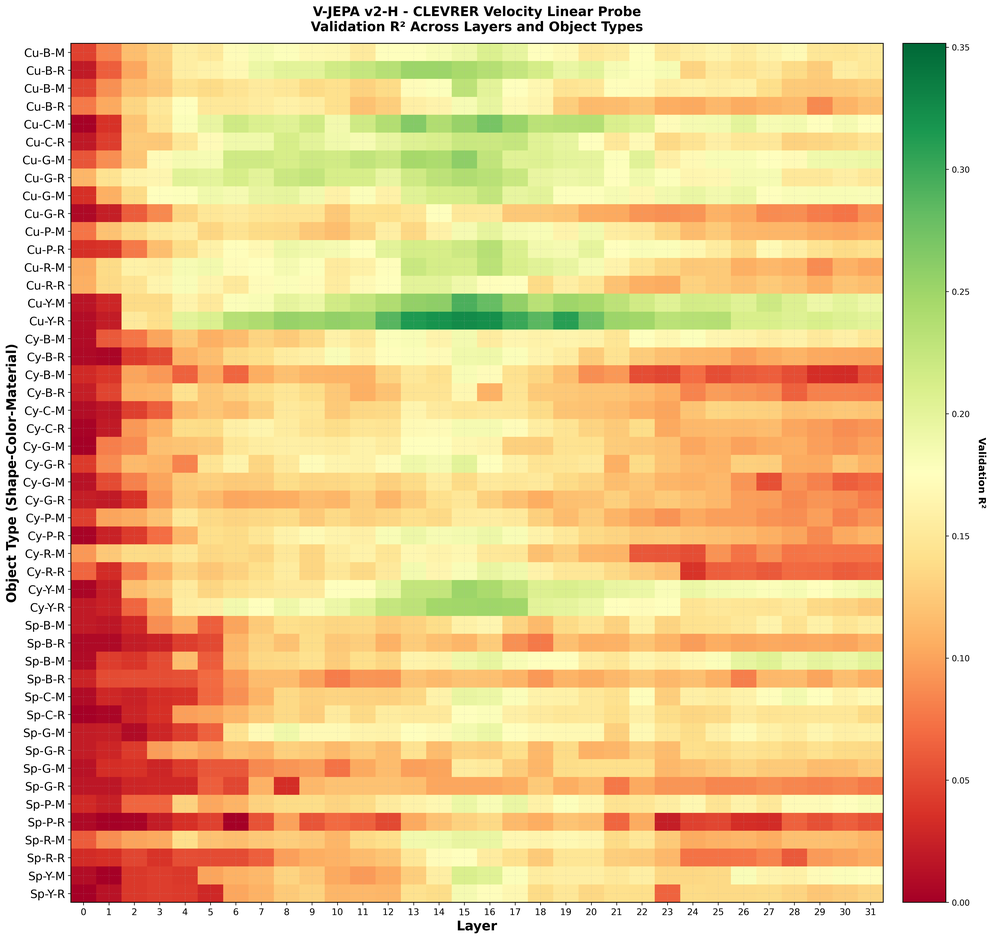}
          \caption{V-JEPA 2-H}
          \label{fig:heatmap_vjepav2h}
      \end{subfigure}
      \hfill
      \begin{subfigure}[b]{0.49\textwidth}
          \centering

  \includegraphics[width=\textwidth]{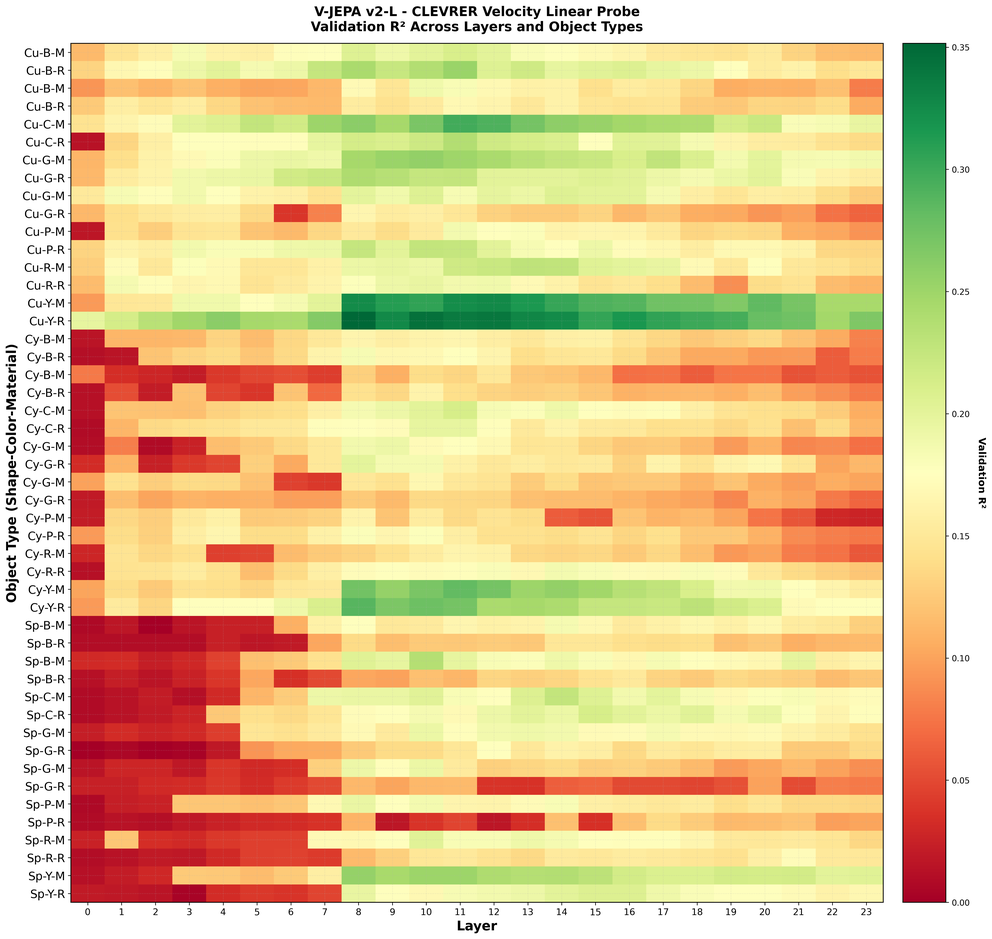}
          \caption{V-JEPA 2-L}
          \label{fig:heatmap_vjepav2l}
      \end{subfigure}

      \caption{CLEVRER velocity linear probe R² heatmaps across layers and object types for all models.}
      \label{fig:heatmap_all_models}
  \end{figure}

\FloatBarrier

\subsection{Subspace Overlap Analysis}                                                                                                                                                     
  \label{app:subspace_analysis}                                                                                                                                               
                                       
  In Section~\ref{sec:orthogonal-intphys-direction}, we examined the relationship between the direction and intuitive physics subtasks to determine whether there is significant             
  representational reuse or shared feature space.                                                                                                                                            
                 
  To quantify the relationship between motion encoding and intuitive physics reasoning, we measure the overlap between the subspaces spanned by direction probes, speed probes, and IntPhys  
  probes across layers.                                                                                                                                                                      
                                                                                                         
  \paragraph{Method.} For each layer $\ell$, we collect the weight vectors from trained probes. For direction (velocity $\theta$) probes, we use circular regression weights                 
  $\mathbf{W}_{\text{dir}} \in \mathbb{R}^{2 \times d}$ (sine and cosine components) from each spatial position, yielding a matrix of probe weights. For speed and IntPhys probes, we collect
   the linear regression weights $\mathbf{w} \in \mathbb{R}^{d}$. We construct orthonormal bases $\mathbf{Q}_A, \mathbf{Q}_B$ for each subspace via QR decomposition of the stacked weight   
  matrices.                                                                                                     
  We compute three metrics to characterize subspace relationships:                                              
  \textit{Principal angles.} The principal angles $\theta_1, \ldots, \theta_k$ between subspaces $A$ and $B$ are computed via the singular value decomposition of $\mathbf{Q}_A^\top         
  \mathbf{Q}_B$, where $\cos(\theta_i) = \sigma_i$ \citep{bjorck1973numerical}. We report the mean principal angle $\bar{\theta} = \frac{1}{k}\sum_{i=1}^k \theta_i$.                        
                                                                                                                       
  \textit{Projection overlap.} The fraction of subspace $B$ captured by subspace $A$ is:                                                                                                     
  \begin{equation}                                                                                                                                                                           
      \text{Overlap}(A \leftarrow B) = \frac{\|\mathbf{Q}_A^\top \mathbf{Q}_B\|_F^2}{\dim(B)}                                                                                                
  \end{equation}                                                                                                                                                                             
  This measures how much of $B$'s variance lies within $A$.                                                                                                                                  
                                 
  \textit{Grassmann distance.} The geodesic distance on the Grassmann manifold:                                                                                                              
  \begin{equation}                                                                                                                                                                           
      d_G(A, B) = \sqrt{\sum_{i=1}^k \theta_i^2}                                                                                                                                             
  \end{equation}                                                                                                                                                                             
                                                                                                                       
  \paragraph{Random Baseline.} To assess whether observed overlaps reflect meaningful structure or merely geometric coincidence, we compare against the expected overlap for random          
  subspaces. For two random subspaces $A$ and $B$ with dimensions $k_A$ and $k_B$ in an ambient space of dimension $d$, the expected projection overlap is \citep{vershynin2018high}:        
  \begin{equation}                                                                                                                                                                           
      \mathbb{E}[\text{Overlap}(A \leftarrow B)] = \frac{k_A}{d}                                                                                                                             
  \end{equation}                                                                                                                                                                             
  Intuitively, a random $k_A$-dimensional subspace captures a fraction $k_A/d$ of any random direction's variance. For V-JEPA 2-L with embedding dimension $d = 1024$:                      
  \begin{itemize}[noitemsep, topsep=0pt]                                                                                                                                                     
      \item \textbf{Direction subspace} ($k = 66$--$136$): expected random overlap = $6.4$--$13.3\%$                                                                                         
      \item \textbf{Speed subspace} ($k = 21$--$29$): expected random overlap = $2.1$--$2.8\%$                                                                                               
      \item \textbf{IntPhys subspace} ($k = 7$--$15$): expected random overlap = $0.7$--$1.5\%$                                                                                              
  \end{itemize}                                                                                                  
  \begin{table}[t]                                                                                                                                              
  \caption{Subspace overlap between motion encoding (direction/speed) and IntPhys probes (V-JEPA v2-L, $d{=}1024$). Overlap$_{A \rightarrow B}$ measures the    
  fraction of subspace $B$ captured by subspace $A$.}                                                                                                           
  \label{tab:subspace_overlap}                                                                                                                                  
  \centering                                                                                                                                                    
  \small                                                                                                                                                        
  \begin{tabular}{@{}lcccccccccc@{}}                                                                                                                            
  \toprule                                                                                                                                                      
  & \multicolumn{5}{c}{Direction vs IntPhys} & \multicolumn{5}{c}{Speed vs IntPhys} \\                                                                          
  \cmidrule(lr){2-6} \cmidrule(lr){7-11}                                                                                                                        
  Layer & Dir & IP & Angle & Dir$\rightarrow$IP & IP$\rightarrow$Dir & Spd & IP & Angle & Spd$\rightarrow$IP & IP$\rightarrow$Spd \\                            
  & Dim & Dim & ($^\circ$) & Overlap & Overlap & Dim & Dim & ($^\circ$) & Overlap & Overlap \\                                                                  
  \midrule                                                                                                                                                      
  0  & 30  & 1  & 79.1 & 0.036 & 0.001 & 25 & 1  & 81.8 & 0.020 & 0.001 \\                                                                                      
  1  & 30  & 1  & 79.6 & 0.033 & 0.001 & 24 & 1  & 82.4 & 0.017 & 0.001 \\                                                                                      
  2  & 14  & 1  & 81.7 & 0.021 & 0.001 & 25 & 1  & 80.3 & 0.028 & 0.001 \\                                                                                      
  3  & 114 & 1  & 70.6 & 0.110 & 0.001 & 17 & 1  & 82.8 & 0.016 & 0.001 \\                                                                                      
  4  & 94  & 1  & 72.8 & 0.088 & 0.001 & 19 & 1  & 82.0 & 0.019 & 0.001 \\                                                                                      
  5  & 62  & 7  & 75.7 & 0.064 & 0.007 & 16 & 7  & 83.5 & 0.015 & 0.007 \\                                                                                      
  6  & 96  & 5  & 71.4 & 0.104 & 0.005 & 20 & 5  & 80.9 & 0.027 & 0.007 \\                                                                                      
  7  & 78  & 11 & 74.6 & 0.073 & 0.010 & 26 & 11 & 81.9 & 0.022 & 0.009 \\                                                                                      
  8  & 136 & 15 & 69.1 & 0.129 & 0.014 & 28 & 15 & 80.7 & 0.030 & 0.016 \\                                                                                      
  9  & 122 & 13 & 70.7 & 0.112 & 0.012 & 21 & 13 & 82.4 & 0.021 & 0.013 \\                                                                                      
  10 & 66  & 12 & 75.2 & 0.069 & 0.012 & 29 & 12 & 81.6 & 0.025 & 0.010 \\                                                                                      
  11 & 120 & 10 & 70.2 & 0.117 & 0.010 & 26 & 10 & 81.2 & 0.027 & 0.010 \\                                                                                      
  12 & 122 & 14 & 69.9 & 0.121 & 0.014 & 22 & 14 & 82.4 & 0.021 & 0.013 \\                                                                                      
  13 & 128 & 13 & 70.0 & 0.119 & 0.012 & 20 & 13 & 82.2 & 0.023 & 0.015 \\                                                                                      
  14 & 100 & 9  & 71.4 & 0.103 & 0.009 & 21 & 9  & 82.6 & 0.018 & 0.008 \\                                                                                      
  15 & 122 & 9  & 70.7 & 0.111 & 0.008 & 26 & 9  & 81.4 & 0.025 & 0.008 \\                                                                                      
  16 & 128 & 14 & 69.7 & 0.122 & 0.013 & 23 & 14 & 81.9 & 0.023 & 0.014 \\                                                                                      
  17 & 128 & 11 & 68.9 & 0.132 & 0.011 & 21 & 11 & 81.9 & 0.023 & 0.012 \\                                                                                      
  18 & 128 & 13 & 69.0 & 0.131 & 0.013 & 24 & 13 & 81.9 & 0.022 & 0.012 \\                                                                                      
  19 & 128 & 7  & 69.0 & 0.130 & 0.007 & 26 & 7  & 81.1 & 0.026 & 0.007 \\                                                                                      
  20 & 400 & 6  & 51.6 & 0.386 & 0.006 & 25 & 6  & 81.3 & 0.026 & 0.006 \\                                                                                      
  21 & 400 & 19 & 51.5 & 0.388 & 0.018 & 26 & 19 & 80.9 & 0.030 & 0.022 \\                                                                                      
  22 & 400 & 6  & 51.6 & 0.386 & 0.006 & 30 & 6  & 79.6 & 0.034 & 0.007 \\                                                                                      
  23 & 400 & 32 & 51.5 & 0.389 & 0.031 & 31 & 32 & 81.3 & 0.030 & 0.031 \\                                                                                      
  \bottomrule                                                                                                                                                   
  \end{tabular}                                                                                                                                                 
  \end{table}                                                  
  \paragraph{Results.} Table~\ref{tab:subspace_overlap} shows the subspace overlap between motion encoding (direction and speed) and IntPhys reasoning across layers. The direction subspace 
  is high-dimensional ($66$--$136$ dimensions), speed is lower-dimensional ($21$--$29$ dimensions), while IntPhys is compact ($7$--$15$ dimensions).                                   
  Direction and IntPhys subspaces maintain mean principal angles of $69^\circ$--$75^\circ$, with direction$\rightarrow$IntPhys overlap of $7$--$13\%$. Speed and IntPhys subspaces are more  
  orthogonal ($80^\circ$--$83^\circ$), with overlap below $3\%$. Critically, these observed overlaps match the random baselines: direction$\rightarrow$IntPhys overlap ($7$--$13\%$) aligns  
  with the expected $6$--$13\%$ for random subspaces of equivalent dimension, and speed$\rightarrow$IntPhys overlap ($2$--$3\%$) matches the expected $2$--$3\%$.                 
  This correspondence with chance-level overlap indicates that the motion and IntPhys subspaces share no more structure than would be expected from arbitrary subspaces of the same          
  dimensionality. Despite both capabilities becoming decodable at similar depths in the network, they occupy nearly orthogonal representational subspaces, ruling out representational reuse 
  and shared latent-variable explanations.                                                 
  \FloatBarrier

\subsection{Direction moves to per-patch encoding}
\label{app:direction-patch-encoding}

\begin{figure*}[h]
   \centering
   \includegraphics[width=0.9\textwidth]{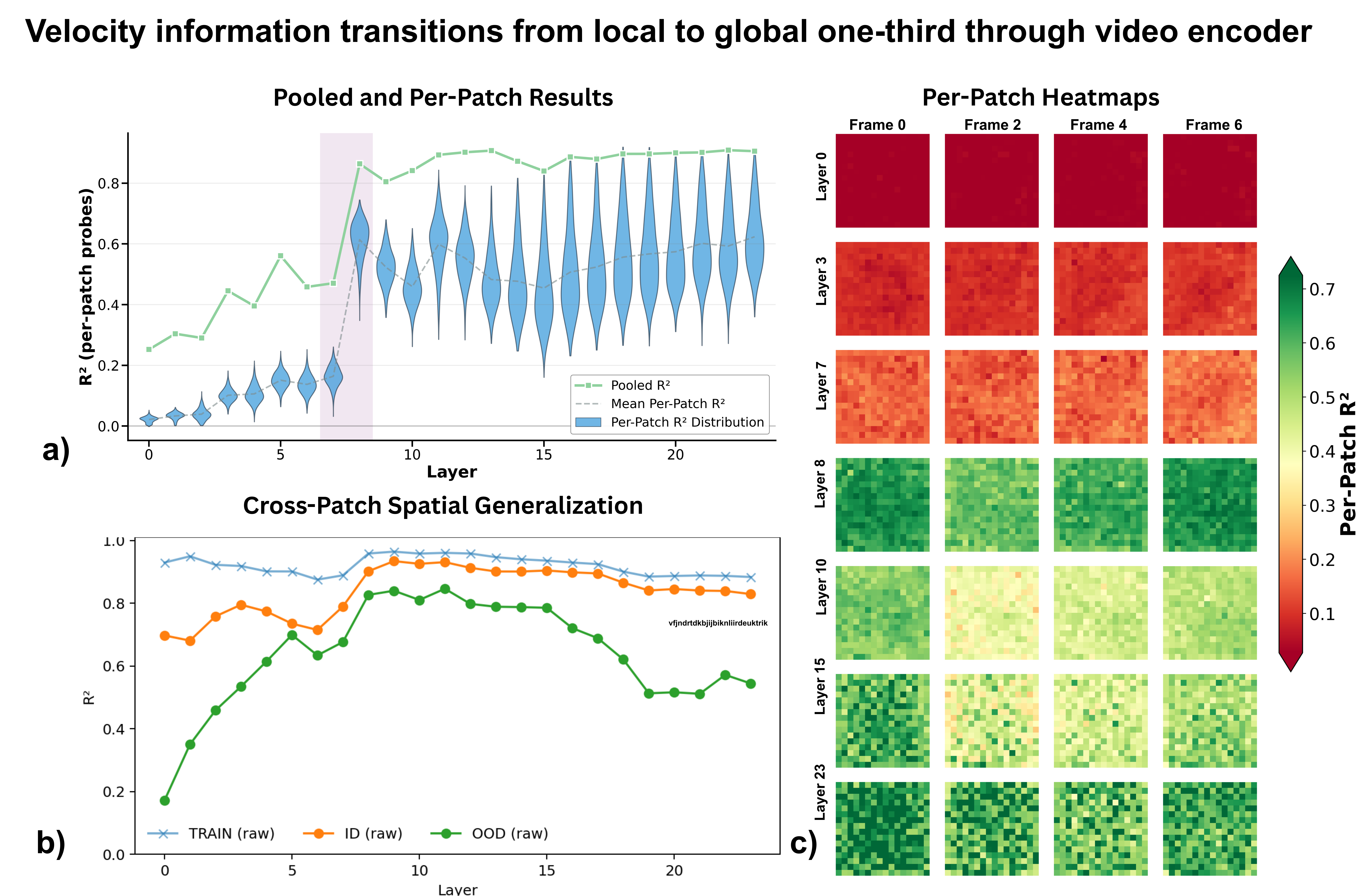}
   \caption{Direction representations transition from local to global at the one-third emergence zone. Results are shown for the synthetic velocity dataset using V-JEPA 2-L. (a) Per-patch linear probe performance across layers. Mean-pooled probes achieve moderate performance earlier by aggregating weak signals across patches, whereas per-patch probes become reliable only at the emergence zone. (b) Spatial generalization performance for probes trained on one half of the frame and evaluated on the other half. Generalization improves sharply at the emergence zone. (c) Per-patch decoding heatmaps illustrate a sharp transition between Layers 7 and 8, where direction becomes decodable from individual patches.}
   \label{fig:per_patch_velocity}
\end{figure*}

In Section~\ref{sec:polar-results}, we showed that direction uniquely arises at the Physics Emergence Zone compared to other physics-related information like scalar speed and scalar acceleration. The one-third emergence zone appears consistently across tasks and datasets, suggesting that it reflects a structural change in the representation rather than a task-specific artifact. We therefore ask what changes in the representation itself when physical information, specifically direction, becomes decodable.

From a patch-level perspective, we find that direction-related signals are already present in early layers, but remain tightly coupled to specific spatial locations. Direction information is fragmented across patches, such that no single patch contains sufficient information to support reliable decoding or spatial generalization. Mean-pooled probes can nevertheless achieve modest performance by linearly combining these fragmented signals across the frame (Figures~\ref{fig:per_patch_velocity}a).

Around the one-third emergence zone, this structure changes sharply. Direction information becomes broadly distributed across patches, with a marked increase in redundancy and spatial spread. At this stage, individual patches begin to carry sufficient information to support reliable direction decoding on their own. This transition explains why per-patch probe performance rises abruptly at the emergence zone, while mean-pooled performance improves more gradually (Figures~\ref{fig:per_patch_velocity}a, b).

This representational shift also enables spatial generalization (Figure~\ref{fig:per_patch_velocity}b). Probes trained on direction information from one region of the frame begin to generalize to unseen regions only after the emergence zone, indicating that direction is no longer tied to specific spatial coordinates. Instead, it becomes encoded in a globally accessible form that survives pooling operations and supports position-invariant decoding.

Together, these results indicate that the emergence zone corresponds to a transition from local, retinotopic direction signals to a globally distributed representation. This local-to-global shift is reminiscent of the V1→MT hierarchy in biological vision, where early motion signals are spatially localized and later representations pool over space to yield position-invariant direction selectivity.

\FloatBarrier

\subsection{Attention Distance Analysis}
\label{app:appendix_attention_distance}

In Section~\ref{sec:local-head-analysis}, we looked at the shared attention head spatiotemporal processing impacted the possible-vs-impossible physics task. We analyze how ablating local attention in V-JEPA v2 affects physics encoding across four tasks: direction prediction ($R^2$), IntPhys accuracy, per-patch direction decoding ($R^2$), and ImageNet classification accuracy. V-JEPA v2 uses tubelet embedding with temporal stride 2, so 16 input frames are encoded into 8 temporal tokens. The model processes $T \times N = 8 \times 196 = 1568$ tokens, where each tubelet covers 2 consecutive frames and $N=196$ is the spatial patch count ($14 \times 14$ grid from $224 \times 224$ images with $16 \times 16$ patches).

We define spatial distance between patches as Euclidean distance in patch coordinates: $d_s(i,j) = \sqrt{(x_i - x_j)^2 + (y_i - y_j)^2}$, measured in patch units (maximum $\sim$18 on the diagonal). Temporal distance is the tubelet difference: $d_t(i,j) = |t_i - t_j|$, ranging from 0 to 7 tubelets.

To test whether local attention is causally important for physics encoding, we ablate attention by masking weights to nearby tokens and renormalizing the remainder. For spatial threshold $s$, we zero all attention where $d_s(q,k) \leq s$; for temporal threshold $t$, we zero attention where $d_t(q,k) \leq t$. We evaluate three regimes: (1) spatial-only ablation with $s \in \{1, 3, 5, 7, 9, 11, 13\}$ patches; (2) temporal-only ablation with $t \in \{1, 2, 3, 4, 5, 6\}$ tubelets; and (3) combined spatiotemporal ablation with paired thresholds.

Large performance drops after ablation indicate reliance on local attention at that scale. Spatial ablation minimally affects direction $R^2$ but degrades per-patch localization; temporal ablation strongly impacts both direction and IntPhys; combined ablation destroys direction encoding entirely.

% \subsection{Attention head locality}

  \begin{figure}[h]
      \centering
      \includegraphics[width=.32\textwidth]{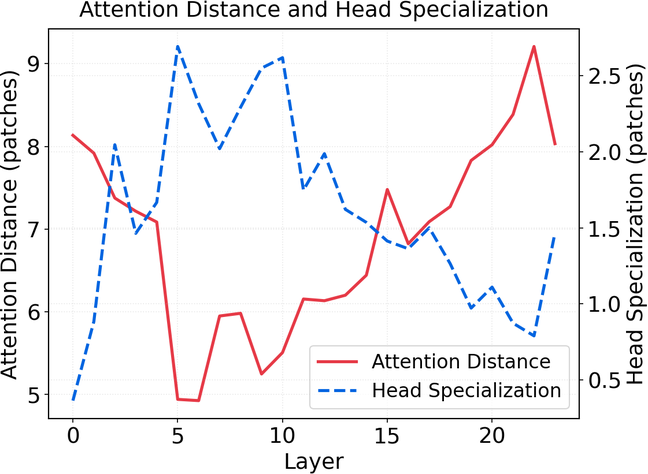}
      \caption{In the physics emergence zone, the average attention head distance drops, and head specialization spikes as local heads emerge among the longer distance heads.}
      \label{fig:attention_locality}
  \end{figure}

\FloatBarrier

% \subsection{Local attention head processing knockout}
% \label{app:full-attention-head-sweep}

\begin{table}[h]
\centering
  \caption{Effect of ablating local attention at varying spatial ($s$, in patches) and temporal ($t$, in tubelets) thresholds.    
  Spatial ablation minimally affects direction $R^2$ but degrades per-patch localization; temporal ablation strongly impacts both 
  direction and IntPhys; combined spatiotemporal ablation destroys direction encoding entirely while IntPhys and ImageNet degrade 
  more gradually.}                                                                                                                
  \label{app:tab:attention-ablation}   
\begin{tabular}{llcccc}                                                                                                                                                                                                                                        
  \toprule                                                                                                                                                                                                                                                                
  & & \textbf{Direction} & \textbf{IntPhys} & \textbf{Per-patch} & \textbf{ImageNet} \\                                                                                                                                                                                                
  \textbf{Condition} & \textbf{Params} & $R^2$ & Acc & $R^2$ & Acc \\                                                                                                                                                                                                                  
  \midrule                                                                                                                                                                                                                                                                             
  Baseline & $s{=}0, t{=}0$ & 0.97 & 78.3 & 0.72 & 33.7 \\                                                                                                                                                                                                                             
  \midrule                                                                                                                                                                                                                                                                             
  \multicolumn{6}{l}{\textit{Spatial-only (temporal preserved)}} \\                                                                                                                                                                                                                    
  & $s{=}1$ & 0.96 & 71.4 & 0.65 & 33.8 \\                                                                                                                                                                                                                                             
  & $s{=}3$ & 0.95 & 67.2 & 0.53 & 33.3 \\                                                                                                                                                                                                                                             
  & $s{=}5$ & 0.95 & 63.9 & 0.43 & 33.5 \\                                                                                                                                                                                                                                             
  & $s{=}7$ & 0.93 & 62.2 & 0.30 & 33.5 \\                                                                                                                                                                                                                                             
  & $s{=}9$ & 0.92 & 60.6 & 0.14 & 33.8 \\                                                                                                                                                                                                                                             
  & $s{=}11$ & 0.91 & 61.1 & $<$0 & 33.7 \\                                                                                                                                                                                                                                            
  & $s{=}13$ & 0.88 & 60.8 & $<$0 & 33.9 \\                                                                                                                                                                                                                                            
  \midrule                                                                                                                                                                                                                                                                             
  \multicolumn{6}{l}{\textit{Temporal-only (spatial preserved)}} \\                                                                                                                                                                                                                    
  & $t{=}1$ & 0.94 & 76.4 & 0.64 & 33.3 \\                                                                                                                                                                                                                                             
  & $t{=}2$ & 0.85 & 60.6 & 0.48 & 33.0 \\                                                                                                                                                                                                                                             
  & $t{=}3$ & 0.83 & 51.9 & 0.41 & 30.3 \\                                                                                                                                                                                                                                             
  & $t{=}4$ & 0.82 & 50.6 & 0.36 & 28.0 \\                                                                                                                                                                                                                                             
  & $t{=}5$ & 0.81 & 50.8 & 0.29 & 27.0 \\                                                                                                                                                                                                                                             
  & $t{=}6$ & 0.80 & 50.8 & 0.24 & 25.6 \\                                                                                                                                                                                                                                             
  \midrule                                                                                                                                                                                                                                                                             
  \multicolumn{6}{l}{\textit{Spatiotemporal (both knocked out)}} \\                                                                                                                                                                                                                    
  & $s{=}3, t{=}1$ & 0.14 & 61.7 & $<$0 & 33.1 \\                                                                                                                                                                                                                                      
  & $s{=}5, t{=}2$ & $<$0 & 60.3 & $<$0 & 31.8 \\                                                                                                                                                                                                                                      
  & $s{=}7, t{=}3$ & $<$0 & 56.4 & $<$0 & 29.7 \\                                                                                                                                                                                                                                      
  & $s{=}9, t{=}4$ & $<$0 & 56.7 & $<$0 & 27.3 \\                                                                                                                                                                                                                                      
  & $s{=}11, t{=}5$ & $<$0 & 58.1 & $<$0 & 19.5 \\                                                                                                                                                                                                                                     
  & $s{=}13, t{=}6$ & $<$0 & 50.8 & $<$0 & 11.2 \\                                                                                                                                                                                                                                     
  \bottomrule                                                                                                                                                                                                                                                                          
  \end{tabular}
  \end{table}

\FloatBarrier

\subsection{Direction Tuning Analysis}                                                                                          
  \label{app:direction_tuning}                                             
  In Section~\ref{sec:direction-ring}, we identified that the Physics Emergence Zone has circular population geometry for direction tuning in tis MLP neurons. To characterize the direction selectivity of individual neurons within V-JEPA 2, we fit generalized linear models (GLMs) to       
  predict each neuron's activation from the stimulus direction. For each neuron $i$ at each spatiotemporal position (patch        
  $\times$ frame), we model the activation $y$ as a linear function of direction $\theta$:                                        
  \begin{equation}                                                                                                                
      y = \beta_0 + \beta_{\cos} \cos(\theta) + \beta_{\sin} \sin(\theta) + \epsilon                                              
  \end{equation}                                                                                                                  
  where $\theta \in [-\pi, \pi]$ is the motion direction in radians. This sinusoidal basis captures smooth, circular tuning curves
   commonly observed in biological direction-selective neurons.                                                                   
              
  We evaluate each neuron's direction tuning strength using cross-validated $\Delta R^2$, computed via $k$-fold cross-validation  
  ($k=5$). For each fold, we fit the GLM on training samples and compute the coefficient of determination on held-out validation  
  samples. The cross-validated $\Delta R^2$ quantifies how much variance in the neuron's response is explained by direction, while
   guarding against overfitting. We regularize the GLM with ridge regression ($\alpha = 10^{-3}$) to ensure numerical stability.  
                                                                                                                      
  To extract each neuron's preferred direction (PD), we fit the GLM on the full dataset and compute:                              
  \begin{equation}                                                                                                                
      \text{PD}_i = \arctan2(\beta_{\sin}, \beta_{\cos})                                                                          
  \end{equation}                                                                                                                  
  This yields the direction that maximally activates each neuron. We also compute the direction tuning gain as                    
  $\sqrt{\beta_{\cos}^2 + \beta_{\sin}^2}$, which reflects the amplitude of the neuron's direction modulation.                    
                                                                                     
  To visualize the population-level direction tuning, we construct heatmaps with neurons on the vertical axis and preferred       
  direction (binned into 24 bins spanning $[-\pi, \pi]$) on the horizontal axis. For each neuron, we assign its cross-validated   
  $\Delta R^2$ to its corresponding preferred direction bin. When multiple spatiotemporal positions for the same neuron fall into 
  the same bin, we aggregate using the maximum $\Delta R^2$ across positions. Neurons are sorted vertically by their peak         
  preferred direction bin, then by tuning strength within each bin. The resulting heatmap reveals whether the network contains a  
  diverse population of direction-tuned neurons spanning all directions, or whether direction tuning is sparse or concentrated at 
  particular angles.   

\begin{figure*}[h]
    \centering
    \includegraphics[width=.8\textwidth]{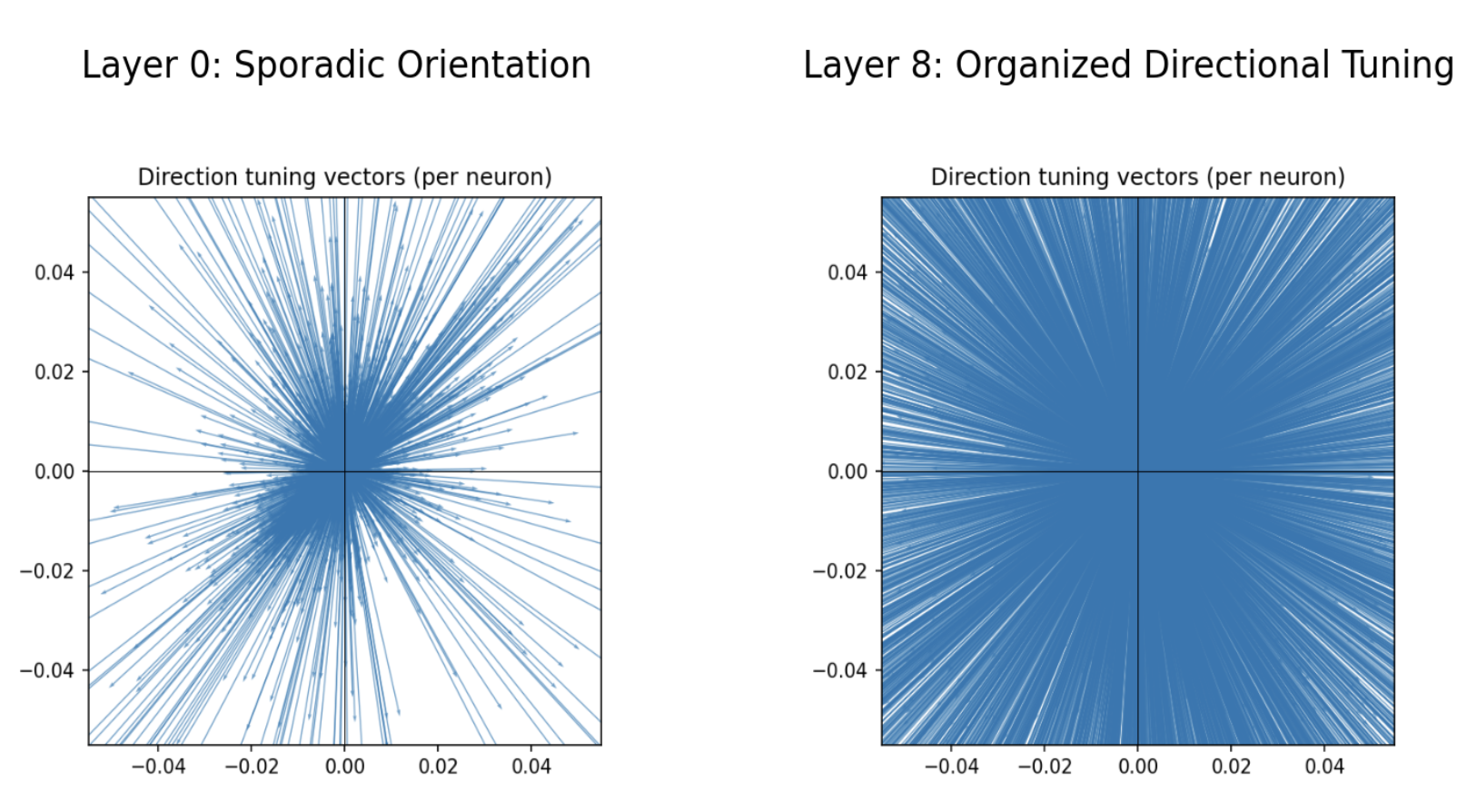}
    \caption{Direction tuning vectors show that direction tuning is sporadic and disorganized at Layer 0, but has emerged in neat organization by Layer 8 of the Physics Emergence Zone}
    \label{fig:layer-0-layer-8}
\end{figure*}

We further find that the circular population code only emerges at the Physics Emergence Zone, and that it is not present at the early layers, with disorganized direction tuning in Layer 0 (Appendix Fig~\ref{fig:layer-0-layer-8})

 \subsection{Speed Tuning Analysis}                                                                                              
  \label{app:speed_tuning}                                                                                                        
    
  To characterize speed selectivity, we fit a separate GLM for each neuron predicting activation from stimulus speed. We use a    
  quadratic model to capture neurons with preferred speeds at intermediate values:                                                
  \begin{equation}                                                                                                                
      y = \beta_0 + \beta_r \cdot r + \beta_{r^2} \cdot r^2 + \epsilon                                                            
  \end{equation}                                                                                                                  
  where $r \geq 0$ is the speed magnitude. The quadratic term allows the model to capture neurons that respond maximally at a     
  particular speed rather than monotonically increasing or decreasing with speed.                                                 
                                                                                                                                  
  As with direction tuning, we evaluate speed tuning strength using cross-validated $\Delta R^2$ with $k$-fold cross-validation   
  ($k=5$) and ridge regularization ($\alpha = 10^{-3}$). To extract each neuron's preferred speed, we compute the vertex of the   
  fitted parabola:                                                                                                                
  \begin{equation}                                                                                                                
      r^*_i = -\frac{\beta_r}{2\beta_{r^2}}                                                                                       
  \end{equation}                                                                                                                  
  This preferred speed is only well-defined when $\beta_{r^2} < 0$ (i.e., the parabola opens downward, indicating a true peak). We
   clip $r^*$ to lie within the observed speed range $[r_{\min}, r_{\max}]$. The speed tuning gain is defined as $|\beta_r|$,     
  reflecting the neuron's sensitivity to speed changes.                                                                           
                                                                                                                                  
  Population heatmaps for speed tuning follow the same construction as for direction: neurons are binned by their preferred speed 
  (24 bins spanning the 1st to 99th percentile of observed preferred speeds), with cross-validated $\Delta R^2$ aggregated via    
  maximum within each bin. Neurons are sorted by their peak preferred speed bin. These heatmaps reveal whether the network        
  contains a continuum of speed-tuned neurons or whether speed encoding is concentrated at particular values.  
    \begin{figure}[h]
      \centering
      \includegraphics[width=0.45\textwidth]{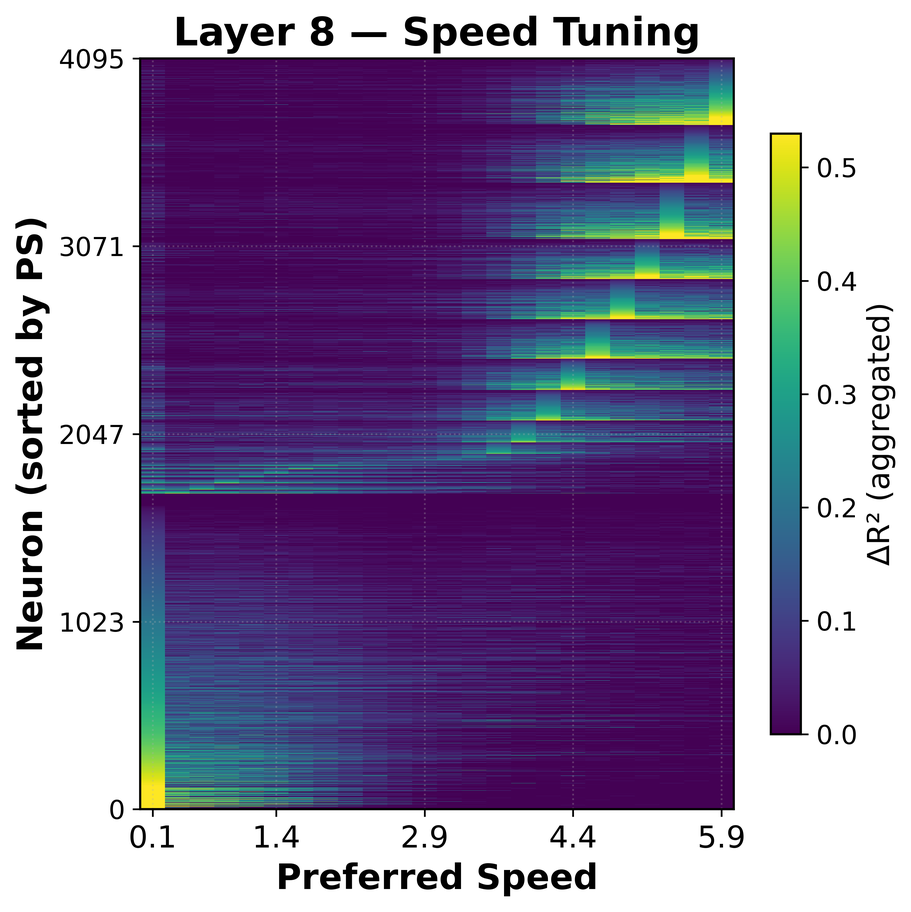}
      \caption{Heatmap of preferred speed for V-JEPA 2 L.}
      \label{app:fig:speed_unit_circle}
  \end{figure}

\FloatBarrier

\subsection{High feature dimensionality of physics-related representations}

In Section~\ref{sec:high-feature-dimensionality}, we discuss the high feature dimensionality of physics-related information.

\begin{figure}[h]
    \centering
    \includegraphics[width=.77\columnwidth]{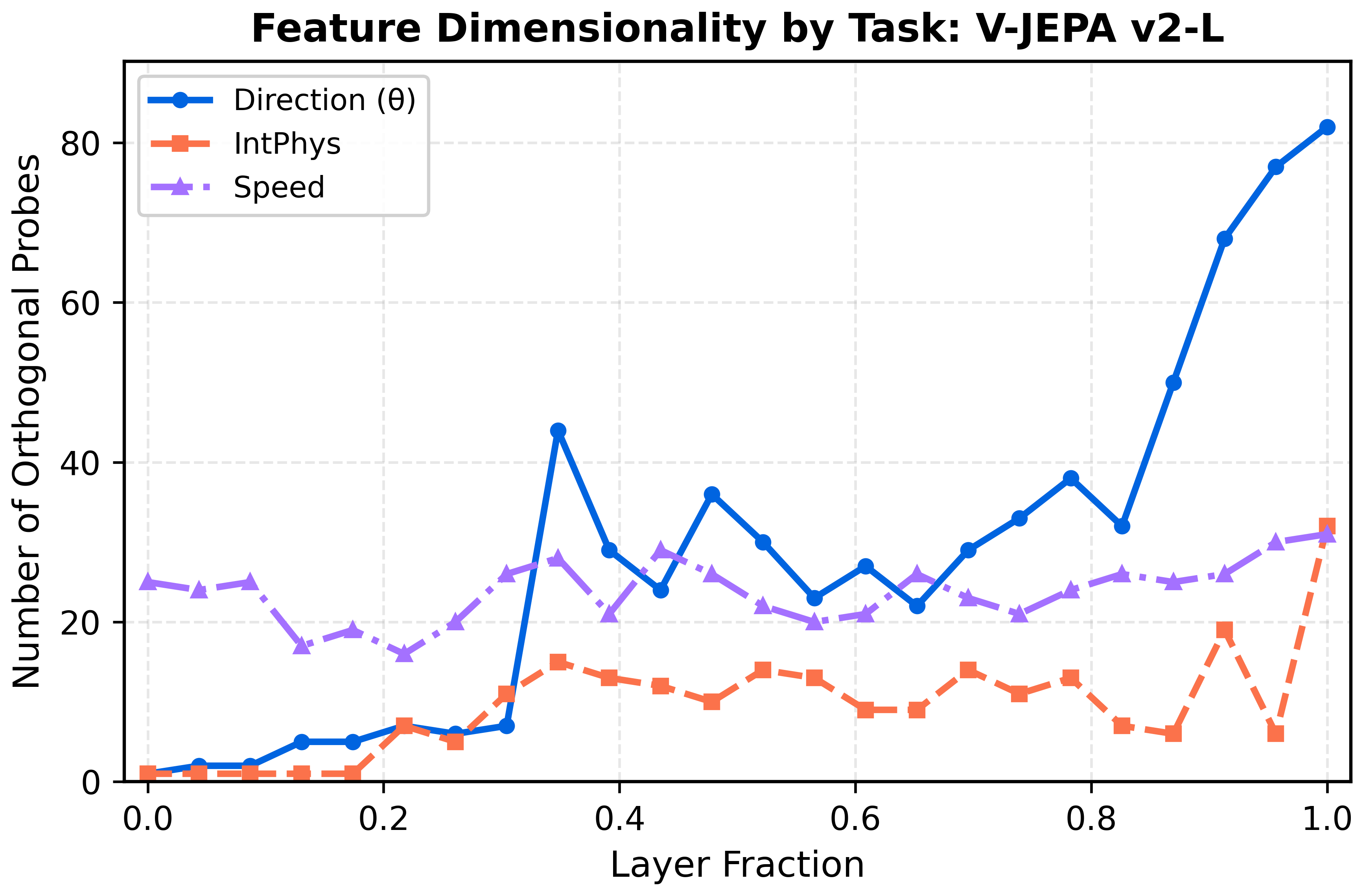}
    \caption{Estimated feature dimensionality of decoded variables across tasks, measured by the number of orthogonal linear probes trainable before performance approaches chance (Direction: $R^2 < 0.3$; Speed: $R^2 < 0.1$; IntPhys: accuracy $< 55\%$). Physical variables require tens of independent features.}
    \label{fig:feature_dimensionality_count}
\end{figure}

\subsection{Speed representations do not have sawtooth pattern}

    \begin{figure}[h]
      \centering
      \includegraphics[width=\textwidth]{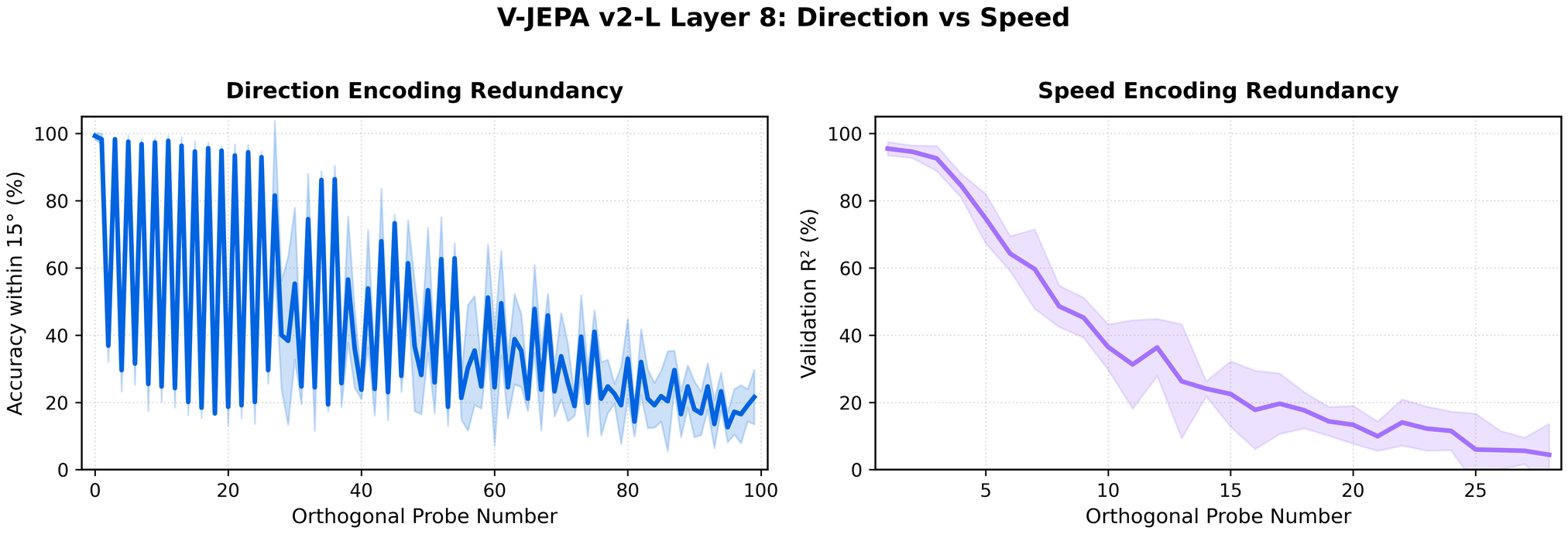}
      \caption{Direction shows a jagged sawtooth pattern, while speed does not.}
      \label{app:fig:speed-sawtooth}
  \end{figure}

\FloatBarrier

\subsection{Orthogonal Probe Sequence Method}      

  \label{app:orthogonal_probes_method}                                                                                                                          
                                                                                                                                                                
  To measure the dimensionality of motion and physics representations, we train sequences of linear probes on progressively orthogonalized activation subspaces.
   This procedure quantifies how many independent directions in activation space encode a given variable.                                                       
                                                                                                                                                                
  \paragraph{Procedure.} For a layer with activations $\mathbf{X} \in \mathbb{R}^{N \times d}$ (where $N$ is the number of samples and $d = 1024$ is the        
  embedding dimension), we iteratively:                                                                                                                         
                                                                                                                                                                
  \begin{enumerate}[noitemsep, topsep=0pt]                                                                                                                      
      \item Train a linear probe $P_k$ on the current activations $\mathbf{X}^{(k)}$ to predict the target variable (direction $\theta$, speed, or IntPhys      
  label).                                                                                                                                                       
      \item Extract the probe weights $\mathbf{W}_k$ and compute an orthonormal basis via QR decomposition.                                                     
      \item Project out the learned direction: $\mathbf{X}^{(k+1)} = \mathbf{X}^{(k)} - \mathbf{X}^{(k)} \mathbf{Q}_k \mathbf{Q}_k^\top$, where $\mathbf{Q}_k$  
  is the orthonormal basis of $\mathbf{W}_k$.                                                                                                                   
      \item Repeat until probe performance falls below threshold.                                                                                               
  \end{enumerate}                                                                                                                                               
                                                                                                                                                                
  The number of probes trained before reaching chance-level performance indicates the effective dimensionality of the representation for that variable.         
                                                                                                                                                                
  \paragraph{Probe Architecture.} All probes are single-layer linear models:                                                                                    
  \begin{itemize}[noitemsep, topsep=0pt]                                                                                                                        
      \item \textbf{Direction ($\theta$):} Circular regression with outputs $(\sin\theta, \cos\theta)$, trained with MSE loss.                                  
      \item \textbf{Speed:} Linear regression with scalar output, trained with MSE loss.                                                                        
      \item \textbf{IntPhys:} Binary logistic regression (possible vs.\ impossible), trained with cross-entropy loss.                                           
  \end{itemize}                                                                                                                                                 
                                                                                                                                                                
  \paragraph{Training Details.} Probes are trained using Adam optimizer with learning rate $\eta = 10^{-3}$ and weight decay $\lambda = 10^{-4}$ for 100 epochs 
  (direction) or 50 epochs (speed, IntPhys). We use an 80/20 train/test split with a fixed random seed for reproducibility.                                     
                                                                                                                                                                
  \paragraph{Stopping Criteria.} The probe sequence terminates when performance approaches chance level:                                                        
  \begin{itemize}[noitemsep, topsep=0pt]                                                                                                                        
      \item \textbf{Direction:} $R^2 < 0.1$ or circular MAE $> 80^\circ$ (chance $\approx 90^\circ$)                                                            
      \item \textbf{Speed:} $R^2 < 0.05$ or MAE $> 90\%$ of random baseline                                                                                     
      \item \textbf{IntPhys:} Accuracy $< 55\%$ or AUC $< 0.55$ (chance $= 50\%$)                                                                               
  \end{itemize}                                                                                                                                                 
                                                                                                                                                                
  \paragraph{Subspace Dimensionality.} The total number of probes $K$ trained before stopping defines the subspace dimensionality. For direction probes (with 2D
   output for $\sin/\cos$), the effective dimensionality is $2K$. For speed and IntPhys probes (1D output), the dimensionality equals $K$. Across layers 0--23, 
  we find direction subspaces of dimension 14--136, speed subspaces of dimension 16--31, and IntPhys subspaces of dimension 1--15                               
  (Table~\ref{tab:subspace_overlap}). 

 \subsection{Steering}                                                                                                                         
  \label{app:steering}                                                                                                                          
                                                                                                                                                
  In Section~\ref{sec:high-feature-dimensionality}, we show that internal representations of physics variables are surprisingly                 
  high-dimensional. Here we verify that the identified direction subspace causally controls direction encoding through activation steering      
  experiments with proper held-out evaluation.                                                                                                  
                                                                                                                                                
  \paragraph{Subspace Construction.} From the orthogonal probe sequence (Appendix~\ref{app:orthogonal_probes_method}), we obtain $K$ probes with
   weights $\mathbf{W}_k \in \mathbb{R}^{2 \times d}$ predicting $[\sin\theta, \cos\theta]$. We construct an orthonormal basis $\mathbf{V} \in  
  \mathbb{R}^{d \times 2K}$ for the direction subspace by stacking the probe weight matrices and applying QR decomposition:                     
  \begin{equation}                                                                                                                              
      \mathbf{V}, \_ = \text{QR}\left([\mathbf{W}_1^\top, \mathbf{W}_2^\top, \ldots, \mathbf{W}_K^\top]\right)                                  
  \end{equation}                                                                                                                                
                                                                                                                                                
  \paragraph{Steering Procedure.} Given activations $\mathbf{x} \in \mathbb{R}^d$ and target angle $\theta^*$, we:                              
  \begin{enumerate}[noitemsep, topsep=0pt]                                                                                                      
      \item \textbf{Project} activations onto the direction subspace:                                                                           
      $\mathbf{c} = \mathbf{V}^\top \mathbf{x}, \quad \mathbf{x}_\perp = \mathbf{x} - \mathbf{V}\mathbf{c}$                                     
                                                                                                                                                
      \item \textbf{Solve} for target coordinates $\mathbf{c}^*$ via least squares such that all probes predict $\theta^*$                      
                                                                                                                                                
      \item \textbf{Reconstruct} steered activations: $\mathbf{x}^* = \mathbf{V}\mathbf{c}^* + \mathbf{x}_\perp$                                
  \end{enumerate}                                                                                                                               
                                                                                                                                                
  \paragraph{Generalization Experiment.} To verify that steering affects the true direction representation---not just the specific probes used  
  for steering---we design a held-out evaluation protocol:                                                                                      
  \begin{enumerate}[noitemsep, topsep=0pt]                                                                                                      
      \item \textbf{Split} the dataset into disjoint train (70\%, 240 videos) and test (30\%, 103 videos) sets                                  
      \item \textbf{Train steering probes} on train set activations using the orthogonal probe sequence (25 probes until $R^2 < 0.1$)           
      \item \textbf{Train a held-out evaluation probe} on test set activations only ($R^2 = 0.99$)                                              
      \item \textbf{Apply steering} (constructed from train-set probes) to test set activations                                                 
      \item \textbf{Evaluate} whether the held-out probe reads the target direction from steered activations                                    
  \end{enumerate}                                                                                                                               
                                                                                                                                                
  This protocol ensures the evaluation probe has never seen the steering probes or the activations used to construct the steering subspace,     
  providing a true test of generalization.                                                                                                      
                                                                                                                                                
  \paragraph{Results.} We steer videos with 8 discrete motion directions ($0°, 45°, 90°, \ldots, 315°$) toward a single target angle $\theta^* =
   90°$. This requires angular shifts ranging from $0°$ (for videos already at $90°$) to $180°$ (for videos at $270°$), with an average shift of
   approximately $90°$. At layer 8 of V-JEPA 2-L:                                                                                              
                                                                                                                                                
  \begin{itemize}[noitemsep, topsep=0pt]                                                                                                        
      \item \textbf{Baseline} (no steering): The held-out probe reads the true direction with MAE $= 4.9°$ to ground truth, but MAE $= 82.9°$ to
   the target angle (as expected given the average angular distance).                                                                           
      \item \textbf{After steering with 20 probes}: The held-out probe reads MAE $= 11.9°$ to the target, demonstrating that steering           
  \emph{generalizes} to a probe trained on entirely different data---a $71°$ improvement over baseline.                                         
  \end{itemize}                                                                                                                                 
                                                                                                                                                
\begin{figure}[h]                                                                                                                             
     \centering                                                                                                                                  
     \includegraphics[width=0.45\textwidth]{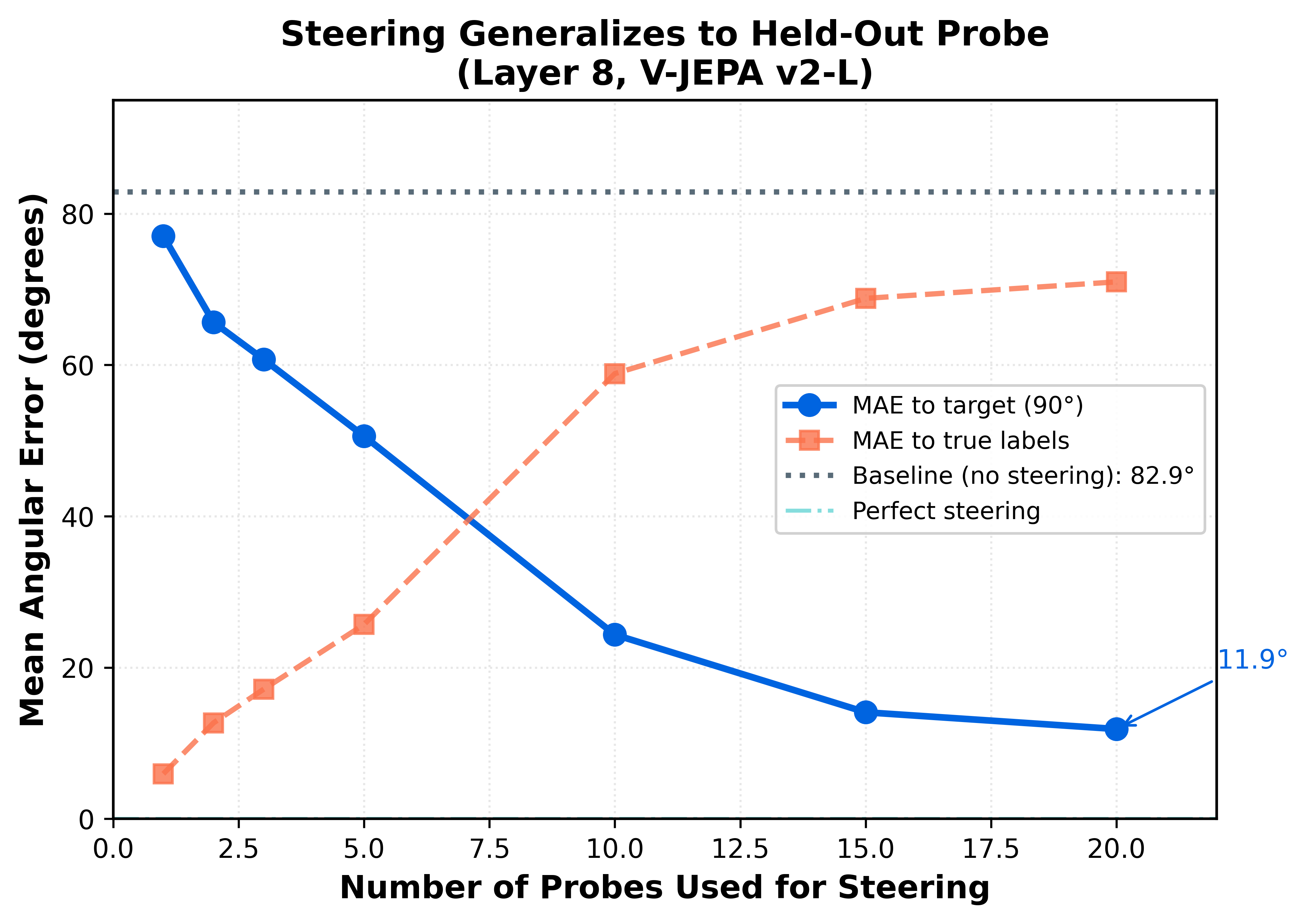}                                         
     \caption{Steering generalizes to held-out data and probes. We train orthogonal probes on the train set (70\%) and a separate evaluation    
  probe on the test set (30\%). Videos span 8 motion directions ($0°$--$315°$), all steered toward target $\theta^* = 90°$. After steering test 
  activations using $N$ train-set probes, the held-out probe (trained only on test data) reads the target direction with MAE decreasing from    
  $82.9°$ (baseline) to $11.9°$ with 20 probes. Layer 8, V-JEPA 2-L.}                                                                          
     \label{fig:steering_generalization}                                                                                                        
  \end{figure}

  Figure~\ref{fig:steering_generalization} shows that effective steering requires coordinated intervention across many orthogonal directions:   
  steering with only 1--5 probes yields modest improvement (MAE $> 50°$), while steering with $\sim$20 probes achieves MAE $\approx 12°$.       
  Importantly, as MAE to the target decreases, MAE to the true labels increases correspondingly, confirming that we are genuinely shifting the  
  representation toward the target direction rather than simply injecting the target into a separate subspace.                                          

\end{document}